%% file: main.tex
\documentclass{article}

\usepackage[top=2.5cm, left=3cm, right=3cm, bottom=3cm]{geometry}
\usepackage{multicol}
\usepackage[english]{babel}
\usepackage[utf8]{inputenc} 
\usepackage[T1]{fontenc}    
\usepackage{hyperref}       
\usepackage{url}            
\usepackage{booktabs}       
\usepackage{amsfonts, mathrsfs, amssymb}       
\usepackage{nicefrac}       
\usepackage{microtype}      
\usepackage[export]{adjustbox}

\usepackage{graphicx}
\usepackage{subfigure,wrapfig}
\usepackage{dsfont}
\usepackage{amsmath}
\usepackage{amsthm}
\usepackage{stmaryrd}
\usepackage{xcolor}

\usepackage{breqn}
\usepackage{shortcuts}
\usepackage{hyperref}
\usepackage{algorithm}
\usepackage{algorithmic}
\usepackage{natbib}
\usepackage[labelfont=bf]{caption}
\usepackage{listings,newtxtt}

\usepackage{etoolbox}
\usepackage{subfiles}
\newtoggle{supplementary}
\toggletrue{supplementary}

\title{\textbf{Wasserstein regularization for sparse multi-task regression}}
\author{Hicham Janati\footnote{INRIA}, \, Marco Cuturi\footnote{Google and CREST / ENSAE}, \, Alexandre Gramfort\footnote{INRIA}}

\date{} 
\begin{document}
\def\subfilebiblio{}
\def\crossref{}
\def\makeannextitle{}

\maketitle

\begin{multicols}{2}

\input{sec/abstract}


\input{sec/introduction}

\input{sec/method}

\input{sec/optim}

\input{sec/experiments}

\input{sec/conclusion}


\bibliographystyle{unsrtnat}

\bibliography{references}
\end{multicols}

\iftoggle{supplementary}{
\cleardoublepage

\subfile{sec/supplementary}

}{}

\end{document}

%% file: sec/abstract.tex
\begin{abstract}
We focus in this paper on high-dimensional regression problems where each regressor can be associated to a location in a physical space, or more generally a generic geometric space. Such problems often employ sparse priors, which promote models using a small subset of regressors. To increase statistical power, the so-called multi-task techniques were proposed, which consist in the simultaneous estimation of several related models. Combined with sparsity assumptions, it lead to models enforcing the active regressors to be shared across models, thanks to, for instance $\ell_{1}/\ell_{q}$ norms.
 We argue in this paper that these techniques fail to leverage the spatial information associated to regressors. Indeed, while sparse priors enforce that only a small subset of variables is used, the assumption that these regressors overlap across all tasks is overly simplistic given the spatial variability observed in real data.
In this paper, we propose a convex regularizer for multi-task regression that encodes a more flexible geometry. Our regularizer is based on unbalanced optimal transport (OT) theory, and can take into account a prior geometric knowledge on the regressor variables, without necessarily requiring overlapping supports.
We derive an efficient algorithm based on a regularized formulation of OT, which iterates through applications of Sinkhorn's algorithm along with coordinate descent iterations. The performance of our model is demonstrated on regular grids with both synthetic and real datasets as well as complex triangulated geometries of the cortex with an application in neuroimaging.
\end{abstract}

%% file: sec/introduction.tex
\section{Introduction}
Several regression problems encountered in the high-dimensional regime involve the prediction of one (or several) values using a very large number of regressors. In many of these problems, these regressors relate to physical locations, describing for instance measurements taken at neighboring locations, or, more generally quantities that are tied by some underlying geometry: In climate science, regressors may correspond to physical measurements (surface temperature, wind velocity) at different locations across the ocean \citep{chatterjee}; In genomics, these regressors map to positions on the genome \citep{laurent}; In functional brain imaging, features correspond to 3D locations in the brain, and a single regression task can correspond to estimating a quantity for a given patient \citep{owen-etal:09}.

These challenging high-dimensional learning problems have been tackled in recent years using a combination of two approaches: \textit{multitask learning} to increase the sample size and \textit{sparsity}. Indeed, it is not uncommon in these problems to aim at predicting several -- not just one -- related target variables simultaneously. When considering \textit{multiple regression tasks}, a natural assumption is that prediction functions (and therefore their parameters) for related tasks should share some similarities. This assumption yields the obvious benefit of being able to pool together different datasets to improve the overall estimation of all parameters~\citep{caruana:93}.
Sparsity has, on the other hand, been a crucial ingredient to help tackle regression problems found for instance in biology or medicine in the ``small n large p'' regime, where the number of observations $n$ is dominated by the dimension $p$ ($n \ll p$). 
For such problems, sparsity-promoting regularizations have lead to important successes, both in practice and theory~\citep{Tibshirani96,Bickel_Ritov_Tsybakov09,Bach_Jenatton_Mairal_Obozinski11}, under the collective name of Lasso-type models.

Challenging problems involving regressors tied by some spatial regularity as those mentioned earlier benefit a lot from the combination of both tools. Indeed, when multiple related regression models in the $p \gg n$ regime need to be estimated, a natural assumption is to consider that each vector of regression coefficients is sparse, and that a common set of active features is shared across all tasks.
This intuition has led to several seminal proposals of Lasso-type models, called multi-task Lasso (MTL) or multi-task feature learning (MTFL)~\citep{argyriou-etal:06,Obozinski06multi-taskfeature}.
Both approaches are based on convex $\ell_{1}/\ell_{2}$ group-Lasso norms that promote block sparse solutions.

An issue alluded to by~\citet{badlinfty} is that perfect overlap between all tasks
can be a too extreme assumption. To understand how to go beyond this binary idea that
active coefficients are the same or not, one can notice that in the context of features mapping to physical locations, employing an $\ell_{1}/\ell_{q}$ norm means
assuming that \textit{exactly} the same locations in the physical space, brain or genome are active for each experiment or patient. This is clearly not realistic in several problems~\citep{gramfort-etal:15}.

\paragraph{Our contribution.}
Our work aims to relax the assumption of perfect overlap across tasks.
To do so, we propose to handle non-overlapping supports in standard multi-task models using an \textit{optimal transport distance} between the parameters of our regression models. Optimal transport (OT) has recently gained considerable popularity in signal processing and machine learning problems. This recent outburst of OT applications can be explained by three factors: the inherent ability of OT theory to compute a meaningful distance between probability measures with non-overlapping supports, faster algorithms to compute that metric using entropic regularization~\citep{cuturi:13}, and their elegant extension to handle non-normalized measures~\citep{chizat:17} at no additional computational cost. Our convex formulation exploits these strengths and applies them to a more general setting in which we consider (signed) vectors. In practice, our regularized problem is optimized using alternating updates, namely fast proximal coordinate descent and Sinkhorn's algorithm.  Sinkhorn iterations are matrix-matrix products which can be sped up on parallel platforms such as GPUs. Our experiments on both synthetic and real data show that our OT model outperforms the state of the art by leveraging the geometrical properties of the regressors.

\paragraph{Related work.}
To extend $\ell_{1}/\ell_{q}$ models and relax full overlap assumption, \citet{dirty} proposed to split the regression coefficients into two parts, one that is common to all tasks and one that is task specific, and to penalize these two parts differently. An $\ell_1$ norm is used to regularize the task-specific part, and an $\ell_{1}/\ell_{q}$ norm is used on the common part. An alternative proposed by \citet{multilevel} is the \emph{multi-level Lasso} (MLL), which considers instead a product decomposition, with $\ell_1$ penalties on both composite variables. Both provide empirical evidence displaying improved performance over block-norm methods. However, experiments show a degraded performance as the overlap between the supports of relevant regressors shrinks.  \citet{tat} propose to learn a tree structure on the features, with inner nodes defined as spatially pooled features. The main advantage of this approach is that no assumptions are made on how tasks are related. However, the inner nodes will be selected if the supports across tasks do not overlap, resulting in spatially smeared coefficients.
Finally, a different approach is proposed by \citet{lobato15} where they consider a sparse multi-task regression with outlier tasks and outlier features (non-overlapping features). They introduce a Bayesian model built on a prior distribution with a set of binary latent variables for each feature and each task.

This paper is organized as follows. Section~\ref{sec:method} introduces our main contribution, the multi-task Wasserstein (MTW) model. We present in Section~\ref{sec:optim} a computationally efficient optimization strategy to tackle the MTW inference problem. Section~\ref{s:experiments} demonstrates with multiple experiments the practical benefits of our model compared to Lasso-type models.

\paragraph{Notation.}
We denote by $\mathds 1_p$ the vector of ones in $\bbR^p$. Given an integer $d \in \bbN$, $\intset{d}$ stands for $\{1, \ldots, d\}$  . The set of vectors in $\bbR^p$ with non-negative (resp. positive) entries is denoted by $ \bbR^p_+$ (resp. $\bbR^p_{++}$).  On matrices, $\log$, $\exp$ and the division operator are applied element-wise. We use $\odot$ for the element-wise multiplication between matrices or vectors. If $X$ is a matrix, $X_{i.}$ denotes its $i^{\text{th}}$ row and $X_{.j}$ its $j^{\text{th}}$ column. We define the Kullback-Leibler (KL) divergence between two positive vectors by $\kl(x, y) = \langle x , \log(x / y) \rangle + \langle y - x, \mathds 1_p \rangle$ with the continuous extensions  $0\log(0 / 0) = 0 $ and $0 \log(0) = 0$. We also use the convention that for $x \neq 0$, $\kl(x | 0) = +\infty$. The entropy of $x \in \bbR^n$ is defined as $E(x) = - \langle x,\log(x) - \mathds 1_p \rangle $. Finally, for any vector $u \in \bbR^p$, the support of $u$ is $\cS_u = \{i \in \intset{p}, u_i \neq 0\}$.

%% file: sec/method.tex

\section{Multi-task Wasserstein model}
\label{sec:method}
\paragraph{Multi-task regression.}
Consider $T$ datasets of labeled vectors $(X^t, Y^t) \in \bbR^{n_t \times p} \times \bbR^{n_t}$,
where $n_t$ is the sample size of each set, and $p$ is the dimension of the common space in which all observations lie.
Our aim is to estimate, in a high-dimensional regime $n_t \ll p$, $T$ linear regression models:
\begin{align*}
    Y^t &= X^t\theta^t + \epsilon^t, \quad t \in \intset{T}\enspace,
\end{align*}
where $\theta^1, \dots, \theta^T \in \bbR^p$ are regression coefficients to be estimated from the
samples $X^t$ with associated labels $Y^t$, and
$\epsilon^1, \dots, \epsilon^T \in \bbR^n$ are additive noise terms assumed to be i.i.d centered Gaussian variables with the same variance $\sigma^2I_n$.
For simplicity, we will assume from now on that $n_t = n$.

\paragraph{Multi-task consensus through Geometric Variance.} 
The idea behind multi-task learning is to estimate $\theta^1,\dots,\theta^T$ jointly, using a regularization term $J$ that promotes some form of similarity between them. 
All multi-task regression models can then be written: 
\begin{equation}
    \label{eq:mtw}
\min_{\theta^1, \dots, \theta^T} \frac{1}{2n} \sum_{t=1}^T{\| X^t \theta^t - Y^t \|_2^2}  +  J(\theta^1, \dots, \theta^T) \enspace. 
\end{equation}
We propose to employ a regularizer that promotes not only sparse solutions, but also some form of ``geometric'' consensus across all $\theta^1, \dots, \theta^T$ through the use of an arbitrary discrepancy function $\Delta: \bbR^p \times \bbR^p \rightarrow \bbR$, writing $
J(\theta^1, \dots, \theta^T) \defeq \min_{\thetabar \in \bbR^p} H(\theta^1, \dots,  \theta^T; \thetabar),
$
where for regularization parameters $\mu \geq 0$ and $ \lambda > 0$,
\begin{equation}
\label{eq:H}
H(\theta^1, \dots,  \theta^T; \thetabar) \defeq \frac{\mu}{T} \!\!\!\!\overbrace{ \sum_{t=1}^{T} \Delta(\theta^t, \thetabar)}^{ \text{geometric variance}} \!\!\!+  \frac{\lambda}{T} \overbrace{ \sum_{t=1}^T \|\theta^t\|_1}^{\text{sparsity}},
\end{equation}
We call the first quantity a geometric variance because it boils down to the usual variance when $\Delta$ is the squared Euclidean distance. Indeed, the minimization of $\bar\theta$ in $J$ would return the mean of all $\theta^t$, and the first sum in $H$ would then be the variance of these vectors.

\paragraph{An OT Discrepancy for Vectors in $\mathbb{R}^p$.} To quantity the geometric variance, we propose to use a new generalized OT metric, that can leverage the fundamental ability of Wasserstein distances to provide a meaningful meta-distance between vectors when a metric on the bins of these vectors is known. However, since OT metrics are defined for positive and normalized vectors, using them in our setting requires some adaptation. Similarly to \citep{sturm, mainini}, we propose to split each vector in its positive and negative parts. 
More formally we write $(x_+, x_-) \in \bbR^p_+$ such that $x = x_+ - x_-$ by setting $x_+ = \max(x, 0)$ applied elementwise. Next, denoting $W$ the unbalanced Wasserstein distance introduced by \citet{chizat:17} and described in detail in the next paragraph, we consider in the rest of this work for two arbitrary vectors $x, y \in \bbR^p$:
\begin{equation}
    \label{eq:extension}
   \Delta(x, y) \defeq W({x}_+, {y}_+)  + W({x}_-, {y}_-) \enspace .
\end{equation}
When $\mu = 0$, \eqref{eq:H} boils down to the penalty of $T$ independent Lasso models, one for each task.
When the $\theta^t$ are fixed, the minimization w.r.t. $\thetabar_+$ (resp. $\thetabar_-$) consists in
estimating the barycenter of the $\theta^t_+$ (resp. $\theta^t_-$) according to the metric $W$. When $\lambda=0$, one forces all the coefficients to be closer according to $W$. 

\paragraph{Unbalanced Wasserstein distance W.}
The reason why optimal transport distances fit our framework is that they can leverage knowledge on the geometry of regressors, in situations such as those presented in the introduction. In OT, that knowledge is known as a \emph{ground metric}. When working in $\bbR^{p}$, this ground metric can be seen as a substitution cost matrix between all $p$ regressors, and is given as a matrix $M \in \bbR_+^{p \times p}$ of pairwise distances between bins. Following the historical analogy of mass displacement cost, $M_{ij}$ represents the cost to move one unit of mass from location $i$ to location $j$. In the current context, $M$ may come from the knowledge that features map to certain spatial positions. For instance, in applications where features correspond to positions ($x_1, \dots, x_p$) in a Euclidean space, the standard cost matrix is given by $M_{ij} = \|x_i - x_j \|_2^2$. 

As proposed in \citep{frogner2015learning,chizat:17}, an optimal transport cost between two nonnegative vectors $\theta_1$ and $\theta_2$ in $\bbR^p_+$ can be defined by seeking a transport plan $P \in \bbR^{p \times p}_+$ that: (i) achieves low transport cost $\langle P, M\rangle$; (ii) has marginals $P\mathds 1$ (resp. $P^\top \mathds 1$) that are as close as possible to $\theta_1$ (resp. $\theta_2$) in KL sense and (iii) has high entropy. These three requirements are reflected in the definition:
\begin{equation}
    \label{eq:uw}
W(\theta_1, \theta_2)\eqdef \min_{\substack{P \in \bbR_+^{p \times p}}} G(P, \theta_1, \theta_2) \enspace,
\end{equation}
where
\begin{equation}
    \label{eq:uw-loss}
        \begin{aligned}
    G(P, \theta_1, \theta_2) = &\overbrace{  \langle P, M\rangle  - \varepsilon E(P)}^{\text{transport - entropy}} + \\ & \overbrace{\gamma \kl(P\mathds 1 | \theta_1) +  \gamma \kl(P^\top \mathds 1| \theta_2) }^{\text{marginal constraints}} \enspace ,
   		\end{aligned}
\end{equation}
and $\varepsilon, \gamma > 0$ are parameters providing a tradeoff between these different objectives.

Large values of $\gamma>0$ tend to strongly penalize unbalanced transports, and as a result penalize discrepancies between the marginals of $P$ and $\theta_1, \theta_2$. The entropy regularization, first introduced by \citet{cuturi:13}, makes the problem strictly convex and computationally faster to solve. A crucial feature of this definition is that the resolution of~\eqref{eq:uw} does not require computing nor storing in memory any optimal plan $P^\star$. Instead, one can study its Fenchel-Rockafellar dual problem given by:
\begin{equation}
\label{eq:uw-dual}
\begin{aligned}
W(\theta_1, \theta_2)= \max_{\substack{u, v \\ \in \bbR_+^p}} & \left[  - \varepsilon \langle u \otimes v  - 1, K \rangle  -  \gamma \langle u^{-\frac{\varepsilon}{\gamma}} - 1, \theta_1 \rangle \right. \\ & \left. - \gamma \langle v^{-\frac{\varepsilon}{\gamma}} - 1, \theta_2 \rangle \right] , 
\end{aligned}
\end{equation}
Performing alternating gradient ascent on~\eqref{eq:uw-dual} amounts to computing matrix scalings of a generalized Sinkhorn algorithm (see Section~\ref{sec:optim}).
\paragraph{Well-posedness.}
We show in this paragraph that a minimizer of \eqref{eq:mtw} exists. To do so, we must prove that the objective function is continuous and coercive.
\begin{Lem}
	\label{lem:w}
	For any $\theta_1, \theta_2 \in \bbR^p_+$ 
	$$W(\theta_1, \boldsymbol0) = W(\boldsymbol0, \theta_2)  = W(\boldsymbol0, \boldsymbol0)  = 0$$
\end{Lem}
\proof 
We show that $W(\boldsymbol0, \theta_2) = 0$.
The result follows directly from the definition of the KL divergence. Let $P \in \bbR_+^{p \times p}$. We have $\kl(P\mathds1, \boldsymbol 0) = 0 $ if $P = \boldsymbol{0}$ and $+\infty$ otherwise. Thus, the minimizer of $G(P,  \boldsymbol0, \theta_2)$ is $P^\star  = \boldsymbol0$ and we have $W(\boldsymbol0, \theta_2) =  G(\boldsymbol0, \theta_1, \bs0) = 0.$  The same reasoning applies to prove $W(\theta_1, \boldsymbol0)  = W(\boldsymbol0, \boldsymbol0) = 0.$ $\blacksquare$

\begin{prop}
	\label{prop:continuity}
	$H$ The extension \eqref{eq:extension} preserves the continuity of $H$ at 0. 
\end{prop}

\proof
Since $H$ is separable across the $(\theta^t)$, we only need to prove that for $\theta, \thetabar \in \bbR_+^{p}$ 
we have $\lim_{(\theta, \thetabar)  \downarrow 0} W(\theta, \thetabar) =\lim_{\thetabar) \downarrow 0} W(0, \thetabar) = \lim_{\theta \downarrow 0} W(\theta, 0) = W(0, 0) = 0$. Let $i, j \in \intset{p}$. Suppose $\theta_i, \thetabar_j \neq 0, 0$. $G$ is smooth, convex and coercive w.r.t to $P$. The first order optimality condition reads:
$$
M + \varepsilon \log(P_{ij}) + \gamma \log\left((P\mathds 1)_i (P^\top \mathds 1)_j \right) = \gamma \log(\theta^t_i \thetabar_j)
$$
$$
\Leftrightarrow \exp(M) (P_{ij})^\varepsilon  \left((P\mathds 1)_i (P^\top \mathds 1)_j \right)^\gamma = (\theta^t_i \thetabar_j)^\gamma
$$
When $(\theta, \thetabar) \to (\boldsymbol0, \boldsymbol0)$, $(P_{ij})^\varepsilon \left((P\mathds 1)_i (P^\top \mathds 1)_j \right)^\gamma  \to 0$ and since $P$ is non-negative we have $\forall i,j, P_{ij} \to 0$, \emph{i.e} $P \to \boldsymbol 0$. The continuity of $G$ with respect to $P$ leads to $\lim_{(\theta, \thetabar)  \downarrow 0} W(\theta, \thetabar) = 0$. Lemma \ref{lem:w} guarantees $\lim_{\thetabar \downarrow 0} W(0, \thetabar) = \lim_{ \theta \downarrow 0} W(\theta, 0) = W(0, 0) = 0$. 
$\blacksquare$

Proposition \ref{prop:continuity} shows that our extension still guarantees that $H$ is continuous at 0. Now we show that the loss function in \eqref{eq:mtw} is coercive.

\begin{prop}
	\label{prop:coercivity}
	The loss function in \eqref{eq:mtw} is coercive.
\end{prop}

\proof 
 Let's prove that $W$ is bounded from below.
 Since $\kl$ is non-negative, and $\langle P, M\rangle \geq 0$, we have $W(\theta_1, \theta_2) \geq \min_{P\in \bbR^{p\times p}}  - \varepsilon E(P) $ which is minimized at $P_{ij}^\star = 1 \, \forall i,j$. Thus~$W(\theta_1, \theta_2)~\geq~-\varepsilon p^2$. Thus, given that the $\ell_1$ norm is non-negative, $H$ is also bounded from below. The coercivity of the loss function follows from the coercivity of the quadratic loss.
 $\blacksquare$

%% file: sec/optim.tex
\section{Efficient Optimization of MTW}\label{sec:optim}
\label{s:optim}%
\paragraph{Loss function.}
We solve MTW by alternating minimization on the positive and negative parts of the regression coefficients $\boldsymbol\theta\defeq (\theta^1,\dots,\theta^T)$ and those of $\thetabar$. We will use in what follows bold symbols for sequences of the form $\boldsymbol{z} = (z^1, \dots, z^T)$.
 Let $\bs{P}_1$ and $\bs{P}_2$ denote respectively the optimal transport plans linking $\bs\theta_+$ with $\thetabar_+$ and $\bs\theta_-$ with $\thetabar_-$, and $\bs{m}_1$ and $\bs{m}_2$ their respective left marginals. Combining \eqref{eq:mtw}, \eqref{eq:H} and \eqref{eq:uw},  the cost function to minimize is given by:
\begin{multline}
\label{eq:mtw-loss}
 L(\btheta; \boldsymbol{P_1}; \boldsymbol{P_2}; \thetabar)  = \sum_{t=1}^T \Big[ \frac{1}{2n} \| X^t \theta^t - Y^t \|^2 + \frac{\lambda}{T} \|\theta^t \|_1 \\ + \frac{\mu}{T}\left[ G(P_1^t, \theta_+^t, \thetabar_+) + G(P_2^t, \theta_-^t, \thetabar_-) \right]\Big] \enspace .
\end{multline}
$L$ is jointly convex in all its variables (since the Kullback-Leibler is jointly convex, proof in Supplementary materials) and the remaining terms are convex and not coupled.
The straightforward solution is to minimize $L$ by block coordinate descent. Since the minimization with respect to the variables $(P_1^t, \theta_+^t, \thetabar_+)_t$ and $(P_2^t, \theta_-^t, \thetabar_-)_t$ is similar, we only detail hereafter the minimization with respect to $(P_1^t, \theta_+^t, \thetabar_+)$. The full optimization strategy is provided in Algorithm~\ref{alg:alt-opt}. 
We alternate with respect to $(\boldsymbol{P_1}, \thetabar_+)$ and each $\theta_+^t$, which can be updated independently and therefore in parallel.
We now detail the two steps of the procedure.
\paragraph{Barycenter update.}
For fixed $\boldsymbol\theta_+$, minimizing with respect to $(\boldsymbol P_1, \thetabar_+)$ boils down to the unbalanced Wasserstein 
barycenter computation of ~\citep{chizat:17} which generalizes previous work by \citet{agueh:11} to compute the minimizer of
$\min_{\thetabar_+ \in \bbR_+^p} \frac{1}{T} \sum_{t=1}^T W(\theta_+^t, \thetabar_+)$. This is equivalent to minimizing simultaneously in $P_1^1, \dots, P_1^t \in \bbR_+^{p \times p}$ and $\thetabar_+ \in \bbR_+^p$ the objective:
\begin{equation}
    \label{eq:ubar}\varepsilon\! \sum_{t=1}^T \kl(P_1^t, K) + \gamma\kl(P_1^t\mathds 1 | \theta_+^t)  +  \gamma \kl({P_1^t}^\top \mathds 1| \thetabar_+) \enspace .
\end{equation}
As pointed out by \citet{chizat:17} and recalled in~\eqref{eq:uw-dual}, Fenchel-Rockafellar duality allows to minimize over dual variables $u^t, v^t \in \bbR^p$ instead of considering plans $P_1^t \in \bbR_+^{p \times p}$. $P_1^t$ can be recovered as $(u^t_i K_{ij} v^t_j)_{ij}$ and its left marginal, needed for the coefficient update, is given by $m_1^t \eqdef P_1^t\mathds 1 = u^t\odot Kv^t$. These steps are summarized in Alg.~\ref{alg:sinkhorn}. We monitor the largest relative change of barycenter the $\thetabar_+$ to stop our loop.

\begin{algorithm}[H]
	\caption{Alternating optimization}
	\label{alg:alt-opt}
	\begin{algorithmic}
		\STATE {\bfseries Input:}  $ \theta^0$, hyperparameters: $\mu, \epsilon , \gamma, \lambda$ and  $M$.
		\STATE {\bfseries Output:} $\btheta$, the minimizer of \eqref{eq:mtw}.
		\REPEAT
		\FOR{$t=1$ {\bfseries to} $T$}
		\STATE Update $\theta_+^t$ with proximal coordinate descent.
		\STATE Update $\theta_-^t$ with proximal coordinate descent.
		\ENDFOR
		\STATE Update the left marginals $m_+^1, \dots, m_+^t$  and $ \thetabar_+$ with generalized Sinkhorn.
		\STATE Update the left marginals $m_-^1, \dots, m_-^t$  and $ \thetabar_-$ with generalized Sinkhorn.
		\UNTIL{convergence}
	\end{algorithmic}
\end{algorithm}

\paragraph{Coefficients update.} 
Minimizing with respect to one $\theta_+^t$ while keeping all other variables fixed to their current estimate yields problem~\eqref{eq:mtw_coef}, where the $\ell_1$ penalty becomes linear due to the positivity constraint. Given the left marginal $m_1$, the problem reads for all $\theta_+^t$ (omitting index $t$):
\begin{equation}
\label{eq:mtw_coef}
\begin{aligned}
\min_{\theta_+ \in \bbR^p_{++} } \Big[&\frac{1}{2n} \|X\theta_+ - X\theta_- - Y\|^2 + \\ & \sum_{i=1}^p \frac{\mu \gamma}{T} ({\theta_+}_i - m_i \log({\theta_+}_i)) + \lambda {\theta_+}_i \Big]\enspace.
\end{aligned}
\end{equation}
The penalty is a separable sum of convex functions with tractable proximal operators, and therefore ~\eqref{eq:mtw_coef} can be solved by proximal coordinate descent~\citep{Tseng01,fercoq}. The following proposition, proved in the appendix, gives a closed-from solution for that proximal operator.
\begin{prop}
	\label{prop:prox} Let $a, b, \alpha \in \bbR_{++}$. Function $g: x \mapsto  (x - a \log(x)) + b x$ is convex on $\bbR_{++}$, and one has:
	\[
	\prox_{\alpha g}(y)\! =\! \frac{1}{2}\left[ - \alpha(b + 1)  + y +\! \sqrt{(\alpha(b + 1) - y)^2 + 4 \alpha a} \right]\enspace .
	\]
\end{prop}

\begin{algorithm}[H]
	\caption{Generalized Sinkhorn\citep{chizat:17}}
	\label{alg:sinkhorn}
	\begin{algorithmic}
		\STATE {\bfseries Input:}  $ \theta^1, \dots, \theta^T$
         \STATE {\bfseries Output:} Wasserstein barycenter of $ \theta^1, \dots, \theta^T$ and marginals $m^1, \dots, m^T$.
		\STATE Initialize for $(t = 1, \dots, T) \, (u^t, v^t) = (\mathds 1, \mathds 1)$, 
		\REPEAT
		\FOR{$t=1$ {\bfseries to} $T$}
		\STATE $u^t \gets \left(\theta^t/Kv^t\right)^{\frac{\gamma}{\gamma + \varepsilon}}$
		\ENDFOR
		\STATE $\thetabar \gets \left( \tfrac{1}{T}\sum_{t=1}^T (K^\top u^t) ^{ \frac{\varepsilon}{\varepsilon + \gamma} }\right)^{\frac{\varepsilon + \gamma}{\varepsilon}} $
		\FOR{$t=1$ {\bfseries to} $T$}
		\STATE $v^t \gets \left(\thetabar/K^\top u^t\right)^{\frac{\gamma}{\gamma + \varepsilon}}$
		\ENDFOR
		\UNTIL{convergence}
 \FOR{$t=1$ {\bfseries to} $T$}
	\STATE $m^t = u^t \odot K v^t$
	\ENDFOR
	\end{algorithmic}
\end{algorithm}
\textbf{Entropy regularization.}
While large values of $\varepsilon$ (strong entropy regularization) induce undesired blurring, low values tend to cause a well-documented numerical instability~\citep{chizat:17, schmitzer16}, which can be avoided by moving to the log-domain~\citep{schmitzer16}. Also for experiments performed on regular grids such as images, one should leverage the separability of the kernel $K$ as proposed in \citep{solomon:15} to recover far more efficient implementations. This also applies to log-domain computations~\citep{schmitz:17}. We use in this work these crucial improvements over naive implementations of Sinkhorn algorithms.

\textbf{Accelerating convergence with warm-start.}
To speed up convergence, we initialize the Sinkhorn scaling vectors to their previous values, kept in memory between two barycenter computations. This does not affect convergence because of the convexity of the objective function. Note that transport plans $P^1, \dots, P^T$ are never instantiated, as this would be too costly. We only compute their left marginals $m^1, \dots, m^T$, which are involved in the coefficients update. We track both the relative evolution of the objective function and that of the norm of the coefficients to terminate the algorithm. Performing less Sinkhorn iterations per barycenter update yields in practice faster convergence, while reaching the same final tolerance threshold. See supplementary materials for an illustration of this tradeoff and a Python implementation of both algorithms.

\textbf{Hyperparameter tuning.}
The MTW model has four hyperparameters: $\epsilon, \gamma, \mu, \lambda$. We provide in this section practical guidelines to set parameters $\varepsilon$ and $\gamma$ within the unbalanced Wasserstein distance.

\textit{Setting $\varepsilon$.}
As mentioned above, entropy regularization speeds up computations but induces blurring. In our experiments we observe that a value of $1/sp$, where $s$ is the median of the ground metric $M$, provides an excellent tradeoff between speed and performance. 

\textit{Setting $\gamma$.}
In the barycenter definition ~\eqref{eq:ubar}, $\gamma$ controls the influence of the marginals: as $\gamma$ goes to  $0$, $P$ tends to $K$ since we can ignore marginal constraints. This transport plan, however, only leads to a local blur with no transport, so that the mass of the barycenter $\thetabar \mathds 1 \to 0.$ To avoid this degenerate behavior, consider the case where $\gamma \gg \varepsilon$ so entropy regularization can be neglected in \eqref{eq:ubar}. The corresponding approximate objective function is given by:
\begin{equation}
    \label{eq:ubar-noreg}
    \sum_{t=1}^T \left[  \langle P^t, M \rangle  +  \gamma \kl(P^t\mathds 1 | a^t)  +  \gamma  \kl(P{^t}^\top \mathds 1| a) \right].
\end{equation}
Deriving the first order conditions, for any $t \in \intset{T}$:
\begin{equation*}
\label{eq:gamma1}
M_{ij} + \gamma \log\left(\frac{P^t_{i.} P^t_{.j}}{a^t_i\bar{a}_j}\right) = 0\quad \text{ and }\quad  \bar{a} = \frac{1}{T}\sum_{t=1}^T {P^t}^\top \mathds 1,
\end{equation*}
By combining the two, we get for any $\tau \in [0, 1]$:
\begin{equation}
    \label{eq:heuristic}
  \gamma \geq -\frac{\max{M}}{\log{\tau}} \Rightarrow  \bar{\psi} \geq \tau \left(\frac{1}{T} \sum_{t=1}^T \sqrt{\psi_t} \right)^2 \enspace ,
\end{equation}
where $\bar{\psi}, \psi_1, \dots, \psi_t$ denote the respective masses of $\bar{a}, a_1, \dots, a^T$ i.e $\psi^t = {a^t}^\top \mathds 1$. Therefore, ~\eqref{eq:heuristic} provides an adaptive parametrization of $\gamma$ that guarantees a lower bound on the mass of $\thetabar$ as a fraction of the $\ell_{0.5}$ pseudo-norm of those of the inputs. In practice, in all experiments we use $\tau = 0.5$ and set $\gamma = \tau \left(\frac{1}{T} \sum_{t=1}^T \sqrt{\psi_t} \right)^2$.

With $\varepsilon$ and $\gamma$ fixed, only two hyperparameters $(\mu, \lambda)$ remain. These control respectively the similarity between tasks and sparsity. Setting two parameters is not more than what is required by Dirty models \citep{dirty} or an Elastic-Net.

%% file: sec/experiments.tex
\section{Experiments}
\label{s:experiments}
\paragraph{Benchmarks.}
To quantify the benefit of multi-task inference,  we use a Lasso estimator independently run on each task as a standard baseline.
We compare the performance of our algorithm against \emph{Dirty models} \citep{dirty} and \emph{Multi-level Lasso} \citep{multilevel}. When the ground truth is available, we evaluate support identification using the area under the curve (AUC) of precision-recall.

The Group Lasso learning model~\citep{argyriou-etal:06,Obozinski06multi-taskfeature} (a.k.a MTFL) can be expressed
by setting the penalty $J$ to be an $\ell_1/\ell_2$ mixed norm: $ \|\boldsymbol\theta\|_{21} = \sum_{j=1}^{p} \sqrt{\sum_{t=1}^T (\theta_j^t)^2}$.
Such a regularization forces all the $\theta^t$ to have the exact same support, $\cS_{\theta^t} = \cS_{\theta^{t'}}$ for all $t,t'$.
To nuance this very strong assumption, Dirty models~\citep{dirty} propose to decompose $\theta^t = \theta_c^t +  \theta_s^t$, where $\theta_c^t$ is common between all tasks (i.e $\cS_{\theta_c^t} = \cS_{\theta_c^{t'}} \forall t, t'$) and $\theta_s^t$ is specific to each one. The regularization then writes:
\begin{equation}
	\label{eq:dirty}
    J_{\text{Dirty}}(\boldsymbol\theta) = \mu \|(\theta_c^1, \dots, \theta_c^T)\|_{21} + \lambda \sum_{t=1}^{T} \|\theta_s^t\|_1 \enspace .
\end{equation}
When $\theta_s = 0$ (resp. $\theta_c = 0$) one falls back to a Group Lasso (resp. independent Lasso) estimator~\citep{argyriou-etal:06,Obozinski06multi-taskfeature}.

Multi-level Lasso (MLL) applies instead the $\ell_1$ penalty on two levels of a product decomposition $\theta_j^t = C_j S_j^t$ where $C \in \bbR^p$ is common across tasks and $S^t \in \bbR^p$ is task specific. For the model to be identifiable, $C$ is constrained to be non-negative. The (MLL) penalty:
\begin{equation}
\label{eq:mll}
J_{\text{MLL}}(S^1, \dots, S^T; C) = \mu \|C\|_1 +  \frac{\lambda}{T} \sum_{t=1}^T \|S^t\|_1 \enspace .
\end{equation}
As shown by \citet{multilevel}, \eqref{eq:mll} is equivalent to a standard multi-task regression problem with the non-convex regularization: 
\begin{equation}
    \label{eq:mll-eq}
J(\boldsymbol\theta)~=~\frac{1}{T}\sum_{j=1}^p \sqrt{\sum_{t=1}^T |\theta^t_j|} \enspace.
\end{equation}

\subsection{Synthetic data}

\begin{figure}[H]
	\includegraphics[width=\linewidth]{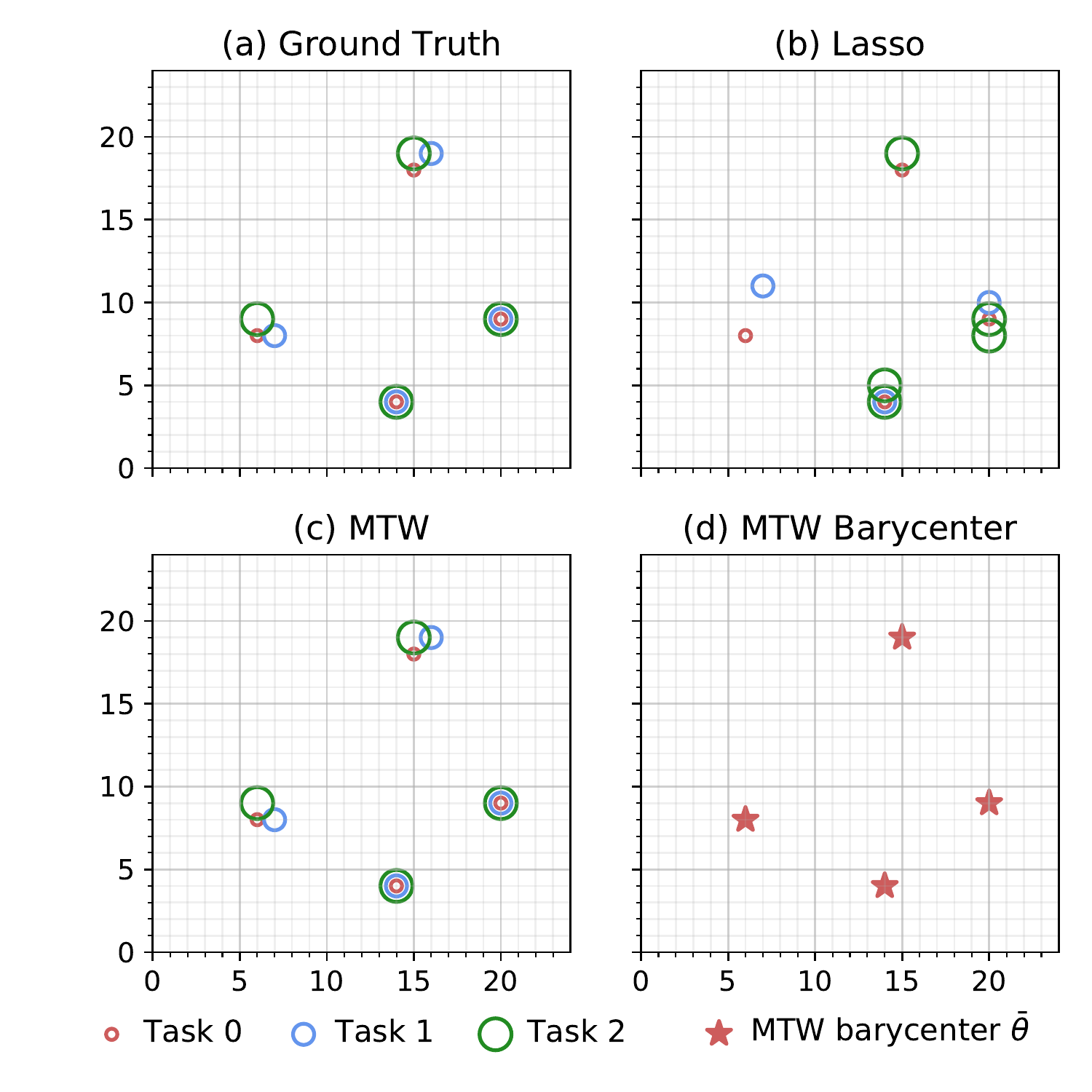}
	\vskip-0.2cm
	\caption{Three sets of color-labeled regression coefficients defined on a 2D grid. Each circle represents a non-zero coefficient. Different radii are used for a better distinction of overlapping features. \textbf{(a)} Inputs. Joint estimation of 3 ill-posed regression tasks using: \textbf{(b)} Lasso \textbf{(c)} MTW model based on a latent Wasserstein barycenter shown in \textbf{(d)}. MTW leverages the proximity of the supports and recovers a closer estimate to the ground truth. The MTW barycenter identifies the 4 key locations across tasks.\label{f:intuition}}\vskip-0.5cm
\end{figure}

 We simulate 3 coefficients $(\theta^t)_{t=1\dots3}$ defined on a 2D grid of shape ($24 \times 24$), and that each vector of coefficients is 4-sparse: each has only 4 non-zero values (see Figure~\ref{f:intuition}). Each coefficient can be seen as a $24 \times 24$ image. We thus have 3 tasks with $p = 576$. The design matrix is obtained by applying a Gaussian filter to the image with standard deviation of 1 pixel, and down-sampling the blurred image by taking the mean over (4 $\times$ 4) blocks. This leads to $n=36$ samples. We set the Gaussian noise variance $\sigma^2$ so that the signal-noise-ratio (SNR) is equal to 3, with $\text{SNR}^2 \eqdef \sum_t \|X^t\theta^t\|_2^2 / (T\sigma^2)$.

\begin{figure}[H]
\centering
\includegraphics[width=\linewidth]{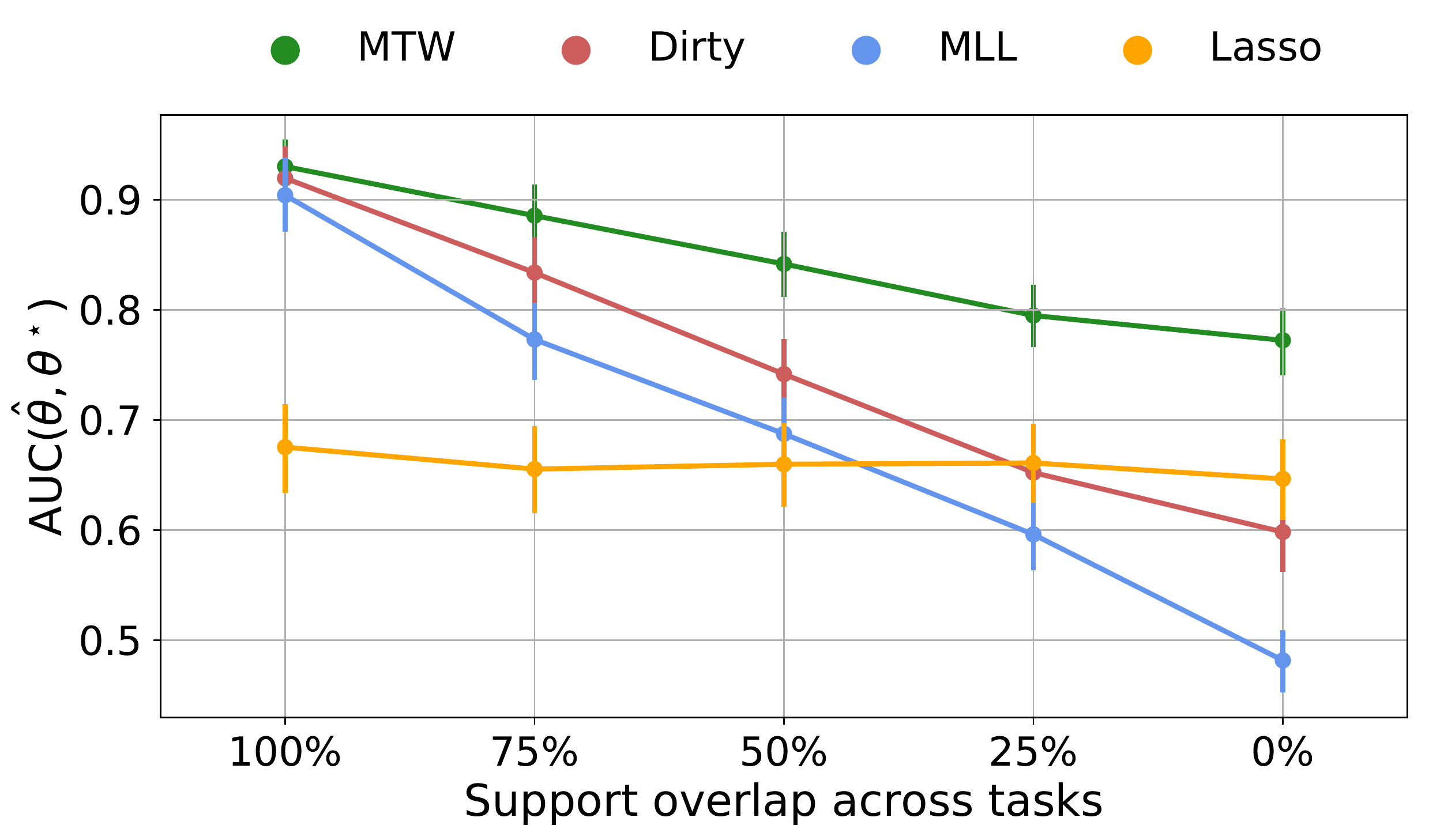}
\caption{AUC scores computed on the estimated coefficients versus ground truth with Dirty models, the Multi-level Lasso (MLL), independent Lasso estimators, and Multi-task Wasserstein (MTW). Mean obtained with 100 independent runs. 
\label{f:auc}}
\end{figure}

To control the overlap ratio between the supports and guarantee their proximity, we first start by selecting two random pixels and randomly translating the non-overlapping features by a one or two pixels for the corresponding tasks. The coefficient values are drawn uniformly between 20 and 30.

Here coefficients map to image pixels, so we employed the MTW with a non-negativity constraint ($\theta^t \in \bbR_+$). 
 Figure \ref{f:auc} shows the distribution of the best AUC scores of 100 experiments (different coefficients and noise). As expected, independent Lasso estimators do not benefit from task relatedness. Yet, they perform better than Dirty and MLL when supports poorly overlap, which confirms the results of~\citet{badlinfty}. MTW however clearly wins in all scenarios.
\subsection{Handwritten digits recognition.}
\begin{figure}[H]
	\centering
	\includegraphics[width=\linewidth]{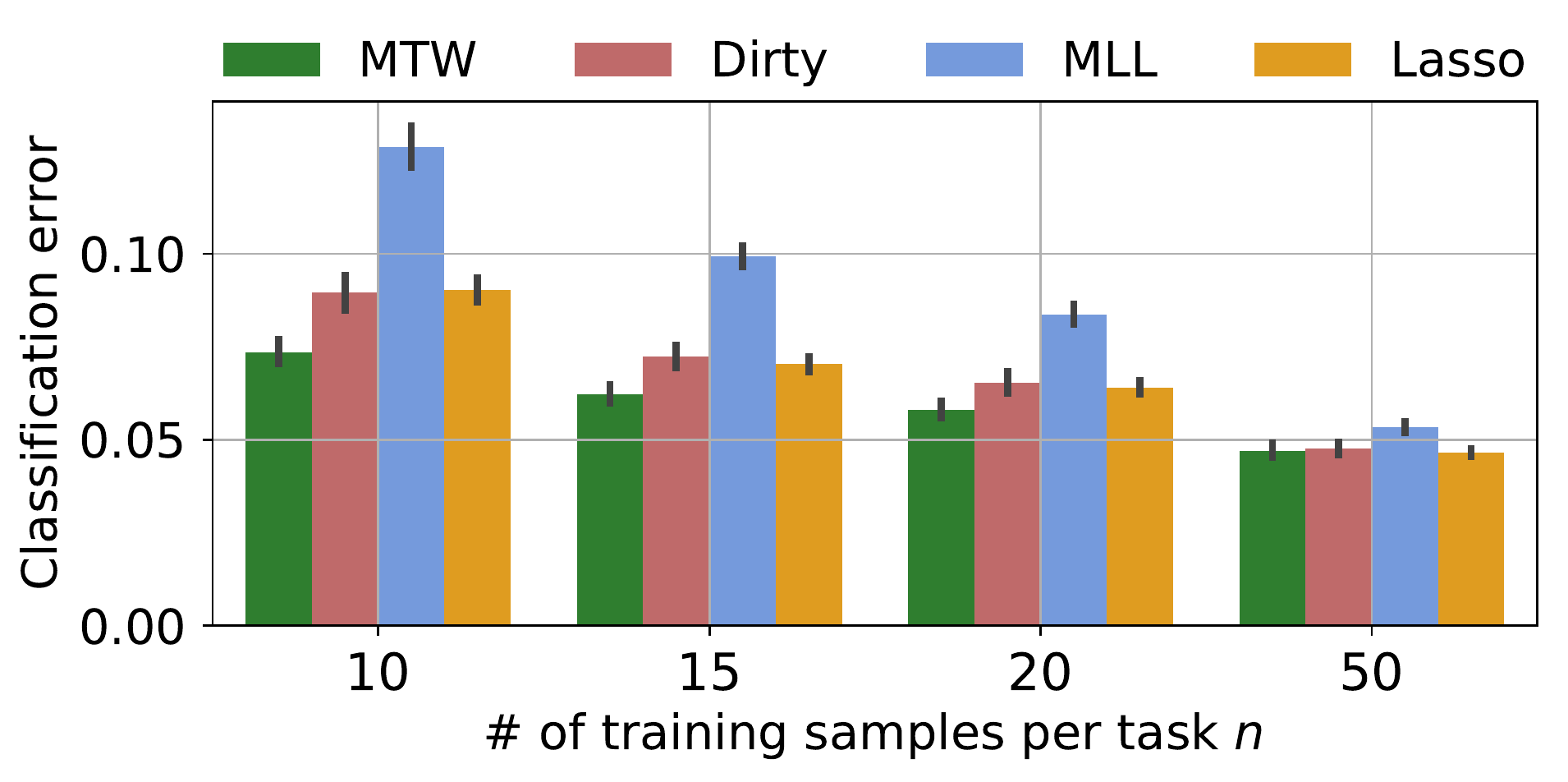}
	\caption{Classification error on a left-out validation set for different numbers of training samples. MTW: Multi-task Wasserstein. MLL: Multi-level Lasso. MTW outperforms all methods as $n$ decreases. Black bars show 95\% confidence intervals over 50 different random splits of the data.
		\label{f:digitsresults}}
\end{figure}

We use the dataset of \cite{digits} consisting of handwritten numerals (`0'--`9') extracted from a collection of Dutch utility maps. 200 patterns per class (for a total of 2,000 patterns) have been digitized in  binary images. We select 6 tasks corresponding to the digits (`0'--`5') and the features corresponding to the pixel averages of (2 $\times$ 3) windows of the original (unprovided) (30 $\times$ 48) handwritten digit images, thus $p=240$. We set $n=$ 10; 15; 20 or 50 training samples per task. Model selection is carried out using a 5-folds cross-validation. We report in figure \ref{f:digitsresults} the mean misclassification rate on the left-out validation set containing $n_v = 200 - n$ samples per task for 50 different random splits of the training / validation data. MTW is particularly efficient in the small $n$ regime with a significant 95\% confidence interval. The regression coefficients obtained by each model are displayed in the appendix.

\subsection{MEG source localization.}
We  use the publicly available dataset DS117 of \citet{ds117}. DS117 contains MEG and EEG recordings of 16 subjects who underwent the same cognitive visual stimulus consisting in pictures of: famous people; scrambled faces; unfamiliar faces. Using the provided MRI scans,  we compute the design matrices $X^t$ i.e the forward operators of the magnetic field generated by a cortical triangulation of $p=2101$ locations using the MNE software~\citep{mne}. The regression outputs $Y^t$ correspond to measurements of the magnetic field on the surface of the scalp recorded by $n=204$ sensors (we keep only MEG gradiometers), as for example used in~\citep{owen-etal:09}. Since the true brain activations $\theta^{\star}$ are unknown, we quantify the performance of our model using the real $(X^t)_{t=1, \dots, T}$ and simulated $(Y^t)_{t=1, \dots, T}$.
Note that the assumption of partial overlap is particularly adapted to this application. Indeed, while functional organization of the brain is comparable between subjects at a certain scale, one cannot assume that the activation foci are perfectly overlapping between individuals. In other words, active brain regions tend to be close in the population but not identical~\citep{thirion-etal:07,xu-etal:09}.

\begin{figure}[H]
 	\centering
 	\adjincludegraphics[trim={0 0cm 0.3cm 0cm},clip, width=\linewidth]{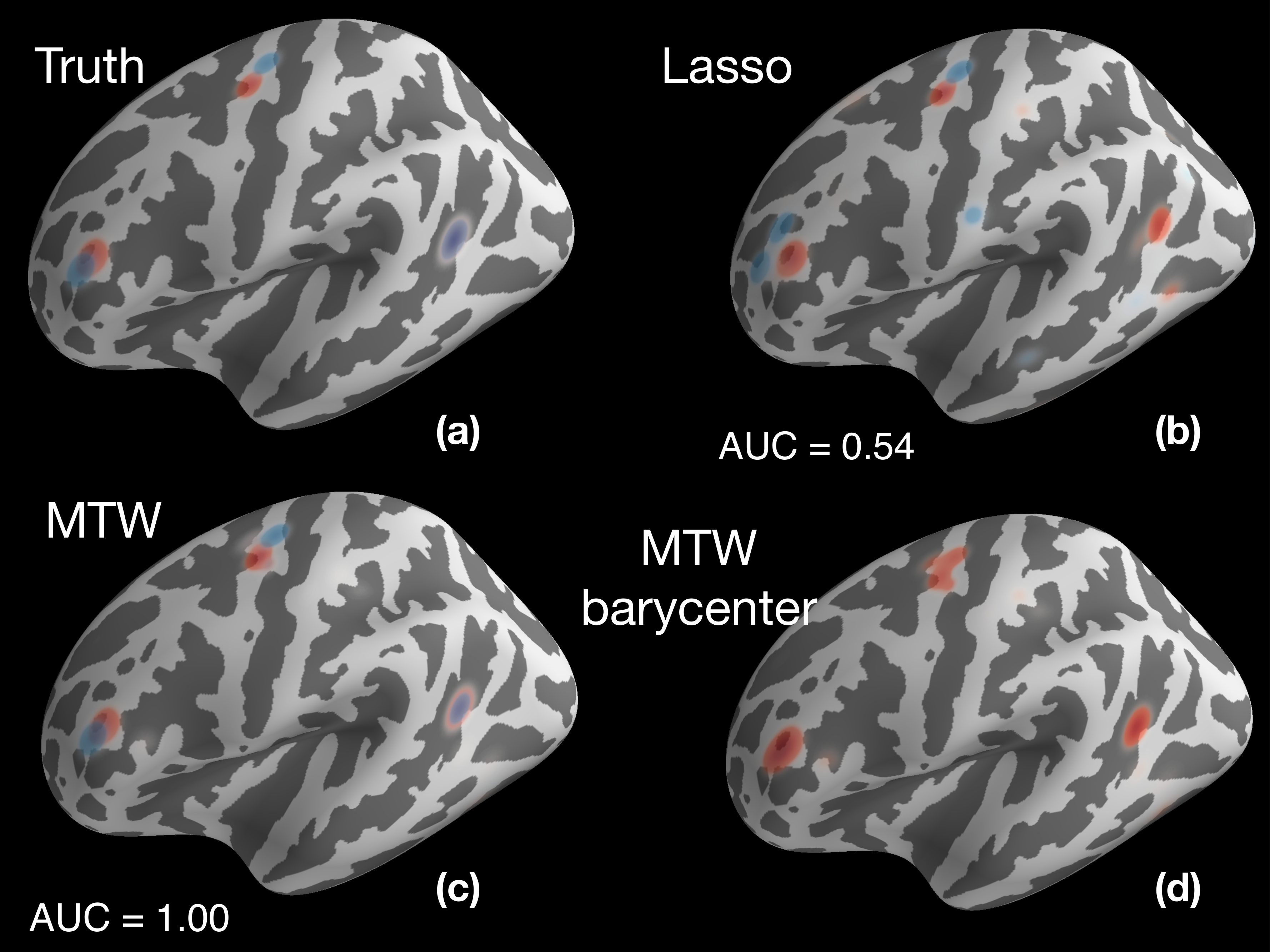}
 	\caption{Activation sources. Each color corresponds to one of the two subjects (except for (d)).  \textbf{(a)}: True sources: one common feature in the back of the brain (right side of the displayed hemisphere) and two non-overlapping sources. \textbf{(b, c)}: Sources estimated by (b) the independent Lasso estimator and (c) the MTW model with the highest AUC score. (d) Shows the barycenter $\thetabar$ associated with MTW model. In this figure, the displayed activations were smoothed for the sake of visibility.
 		\label{f:meg}}
 \end{figure}

\begin{figure}[H]
\centering
		\includegraphics[width=\linewidth]{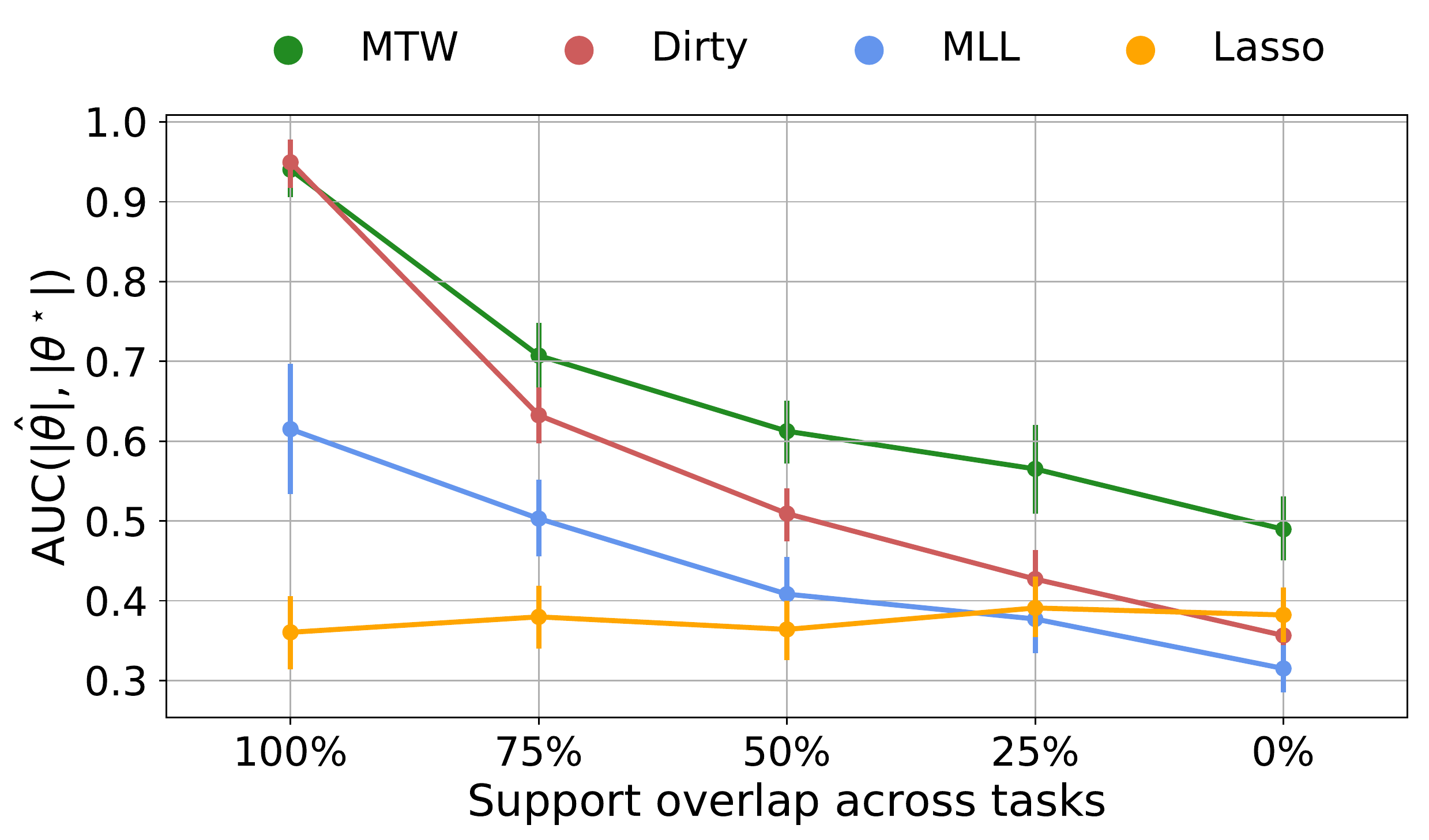}
\caption{Comparaison of different values of supports overlap using 3 tasks. Mean AUC score obtained on MEG data simulation with a SNR = 4 over 30 different experiments.
MTW: Multi-task Wasserstein. MLL: Multi-level Lasso. MTW outperforms other models for all supports overlap fractions.
	\label{f:brains-auc}}
\end{figure}

\paragraph{Simulated activations.}
The regression coefficients (sources) are $k$-sparse ($k \in \intset{11}$), \emph{i.e} all zero except for $k$ random locations chosen respectively in one of 11 distinct brain regions (displayed in supplementary material). Their amplitudes are taken uniformly within $20-30$\,nAm. Their sign is then decided by a coin toss (Bernoulli with 0.5 parameter).  We generate in this manner a set of different regression coefficients for the number of tasks desired. We construct the outcome $Y^t$ with a SNR equal to 4. For MTW, the ground metric $M$ is the distance on the cortical mesh of $p$ vertices. It corresponds to the geodesic distance on the complex topology of the cortex.
\paragraph{Illustrative example.}
MTW is expected to be most valuable for non-overlapping supports. To visually illustrate the benefits of our model, we randomly select 2 subjects and simulate regression coefficients with 3 sources per task with only one common feature. Figure~\ref{f:meg} shows MTW at its best: MTW leverages the geometrical proximity of the sources and thereby perfectly recovers the true supports. The independent Lasso estimator however reaches a poor AUC of 0.54. It selects features very far from the true brain regions which can lead to erroneous conclusions. Moreover, the latent barycenter $\thetabar$ highlights the most representative sources of the cohort of subjects studied (Fig.~\ref{f:meg} (d)). 
\paragraph{Effect of degree of overlap.}
 Using 3 subjects, we perform 30 trials with different noise and coefficients locations and values (Figure~\ref{f:brains-auc}). We make the localization even harder by selecting 5 sources, \emph{i.e.} 5 non-zero features per task. We select the best performance of all models in terms of AUC score. MTW outperforms all benchmarks in recovering the true supports in all scenarios. Unlike Dirty models, MLL fails to recover perfectly overlapping supports and has a large variance. This behavior may be due to the non-convexity of the penalty in \eqref{eq:mll-eq} and potentially bad local minima.
 
\begin{figure}[H]
	\includegraphics[width=\linewidth]{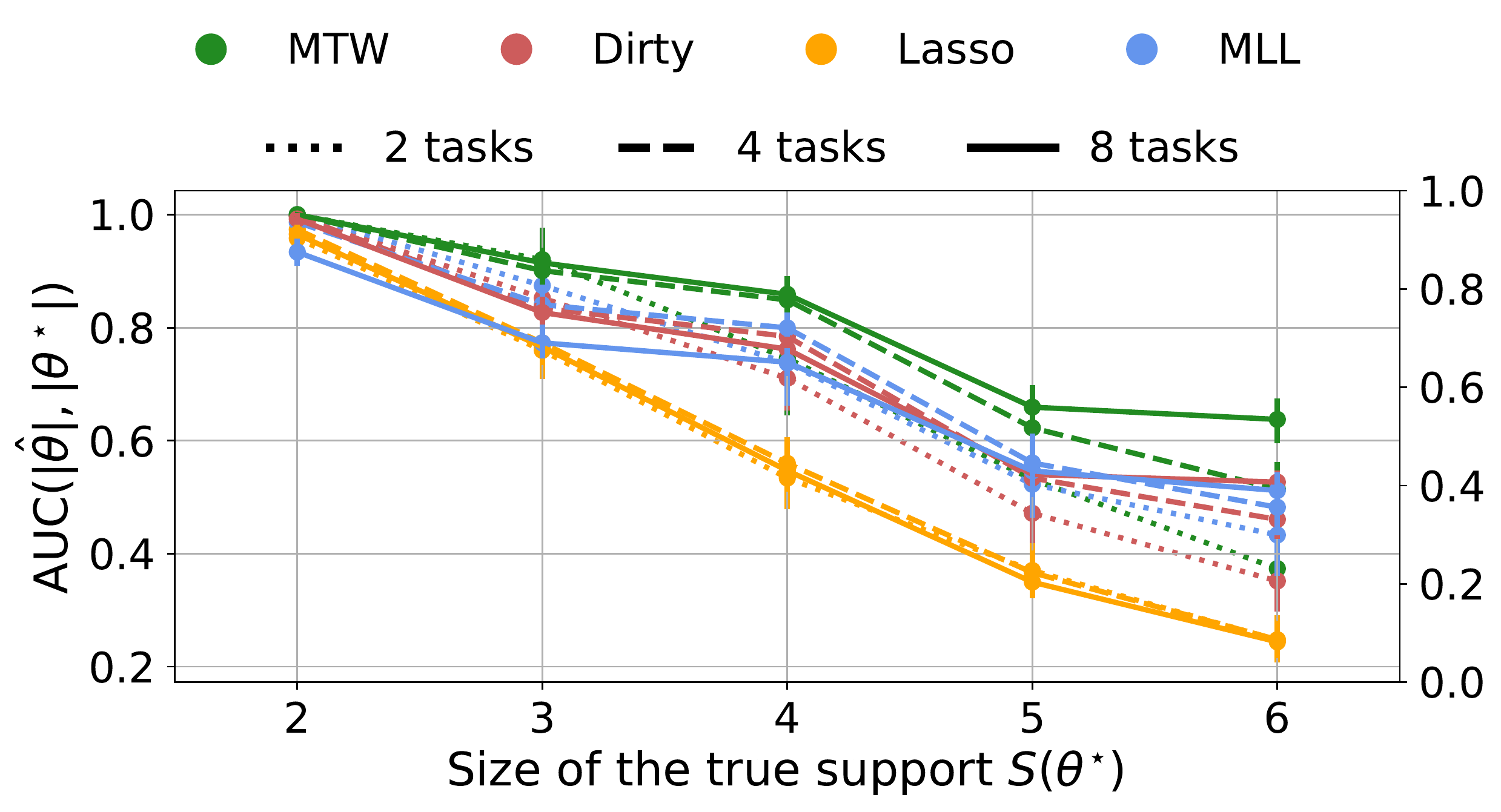}
	\caption{Mean AUC score for different numbers of tasks and support sizes with an overlap of 50\% over 20 different experiments. MTW benefits more from additional learning tasks; obtained on MEG data simulation with a SNR = 4. 
		\label{f:ntasks-auc}}
\end{figure}

\paragraph{Effect of number of tasks.}
When the number of non-zero features increases, recovering the support is more difficult.  Figure \ref{f:ntasks-auc} shows that MTW handles particularly well that scenario, as tasks increase. We compute the mean AUC score of 20 trials for 2, 4 and 8 tasks, 2 to 6 non-zero coefficients with an overlap of supports set to 50\% and a SNR equal to 4. The curves obtained by independent Lasso overlap as it does not benefit from additional tasks. Dirty models handle relatedness through the $\ell_1/\ell_2$ penalty which only improves the estimation of the common features across tasks. This explains why the performance of Dirty models with 4 and 8 tasks is the same. MTW is unique in that it benefits from all 8 tasks.

%% file: sec/conclusion.tex
\section*{Conclusion}
The seminal work of~\citet{caruana:93} has motivated a series of contributions leveraging the presence of multiple and related learning tasks (MTL) to improve statistical performance. Our work is one of them in the context of sparse high dimensionial regression tasks where regressors can be associated to a geometric space. Using Optimal Transport to model proximity between coefficients, we proposed a convex formulation of MTL that does not require any overlap between the supports, contrarily to previous literature. We show how our Multi-task Wasserstein (MTW) model can be solved efficiently relying on proximal coordinate descent and Sinkhorn's algorithm. Our experiments on synthetic and real data demonstrate that regardless of overlap, MTW leverages the geometry of the problem to outperform standard multi-task regression models.

\section*{Acknowledgments}
MC acknowledges the support of a \emph{chaire d'excellence de l’IDEX Paris Saclay}. AG was supported by the European Research Council Starting Grant SLAB ERC-YStG-676943.

%% file: sec/supplementary.tex
\makeannextitle

\appendix
\numberwithin{figure}{section}
\numberwithin{equation}{section}

This appendix is organized as follows. Section~\ref{s:suppmtw} presents details on MTW: convexity, proximal coordinate descent and some background on Sinkhorn's algorithm where we discuss a log-stabilized version~\cite{schmitzer16} that is used in all our experiments. Section~\ref{s:dirty} provides mathematical details on tuning the hyperparameters of Dirty models. Section~\ref{s:expes} provides further details on model selection and experiments. Finally, section~\ref{s:code} provides the Python code used in our experiments.

\section{Technical details on MTW}
\label{s:suppmtw}
\paragraph{Joint convexiy.}
Recall the loss function:

\begin{align*}
 L(\btheta; \boldsymbol{P_1}; \boldsymbol{P_2}; \thetabar) & = \sum_{t=1}^T \Big[ \frac{1}{2n} \| X^t \theta^t - Y^t \|^2 + \frac{\lambda}{T} \|\theta^t \|_1 \\ + &\frac{\mu}{T}\left[ G(P_1^t, \theta_+^t, \thetabar_+) + G(P_2^t, \theta_-^t, \thetabar_-) \right]\Big] \enspace .
\end{align*}

where 

\begin{align*}
G(P, \theta_1, \theta_2) = &\overbrace{  \langle P, M\rangle  - \varepsilon E(P)}^{\text{transport - entropy}} + \\ & \overbrace{\gamma \kl(P\mathds 1 | \theta_1) +  \gamma \kl(P^\top \mathds 1| \theta_2) }^{\text{marginal constraints}} \enspace ,
\end{align*}
   		
The quadratic loss function and the $\ell_1$ penalty are convex and separable across the $(\theta^t)_t$. The transport and entropy terms in $G$ are convex and separable across the $(P^t)_t$. The only coupled terms involved in $L$ are the marginal constraints in $G$. To prove joint convexity of $L$ we only need to prove that of KL (since taking out the marginal is a linear operator). 

Let $x, y \in \bbR_+^p$. We defined the Kullback-Leibler function as: 
\[
\kl(x, y) = \sum_{i=1}^p x_i\log(x_i / y_i) + y_i - xi
\]
Since KL is an element-wise sum, all we need to show is the joint convexity of $f: (a, b) \mapsto a\log(a / b)$ in $\bbR_+^2$.

Let $\tau \in [0, 1]$ and $a_1, a_2, b_1, b_2 > 0$.
Denote $a_{\tau} = \tau a_1 + (1-\tau)a_2$ and $b_{\tau} = \tau b_1 + (1-\tau)b_2$. And let $g: x \mapsto x\log(x)$.

$g$ is convex. Using Jensen's inequality:
\begin{align*}
    f(a_{\tau}, b_{\tau}) &= a_{\tau} \log(a_{\tau} / b_{\tau}) \\
    &= b_{\tau} g(a_{\tau} / b_{\tau}) \\
    &= b_{\tau} g\left(\frac{\tau b_1}{b_{\tau}} \frac{\tau a_1}{\tau b_1} + \frac{(1-\tau) b_2}{b_{\tau}} \frac{(1-\tau) a_2}{(1-\tau)b_2}\right) \\
    &\leq b_{\tau}  \left( 
    \frac{\tau b_1}{b_{\tau}} g\left(\frac{\tau a_1}{\tau b_1}\right) + \frac{(1-\tau) b_2}{b_{\tau}} g\left(\frac{(1-\tau) a_2}{(1-\tau)b_2}\right)
    \right) \\
    &= \tau b_1 g\left(\frac{a_1}{b_1}\right) + (1-\tau) b_2 g\left(\frac{a_2}{b_2}\right)\\
    &=  \tau f(a_1, b_1) + (1-\tau) f(a_2, b_2)
\end{align*}
Therefore, $f$ is jointly convex. $\square$

\paragraph{Coordinate descent.}
Recall that the optimization problem solved by our estimator MTW is carried out by alternating between independent coefficients updates and a barycenter computation. First, we give a proof for Proposition 3.1 just recall here:

\begin{prop}
    \label{prop:prox} Let $a, b \in \bbR_+$. The function $g: x \mapsto  (x - a \log(x)) + b x$ is convex and proximable on $\bbR_{++}$, moreover its proximal operator is given by:
    \[
    \prox_{\alpha g}(y) = \frac{1}{2}\left[ - \alpha(b + 1)  + y + \sqrt{(\alpha(b + 1) - y)^2 + 4 \alpha a} \right]\enspace .
    \]
\end{prop}

\emph{Proof.} 
$g$ is clearly convex. Its proximal operator, defined on $\bbR_{++}$, is given by the minimizer of the problem:
\begin{align*}
    \prox_{\alpha g}(y) &= \min_{x} \frac{1}{2} (x - y)^2 + \alpha g(x) \\
                     &=  \min_{x} \frac{1}{2} (x - y)^2 +   - \alpha a \log(x) + \alpha (b + 1) x
\end{align*} 
The objective function above is differentiable, strictly convex and goes to $+\infty$ when $x \to 0^+$ or $x \to +\infty$. Thus, its minimizer is unique and is the solution of the necessary first order optimality condition: 
\begin{align*}
x - y - \frac{\alpha a}{x} + \alpha b + \alpha = 0 \\ 
\Rightarrow x^2 + \alpha(b + 1) - y x - \alpha a = 0
\end{align*}

The positive solution of the quadratic equation above is given by $x = \frac{1}{2}\left[ - \alpha(b + 1)  + y + \sqrt{(\alpha(b + 1) - y)^2 + 4 \alpha a} \right] $.  \qed
 
 Now recall the coefficient update problem:

\begin{equation}
\label{eq:mtw_coef2}
\min_{\theta \in \bbR^p_{++} } \frac{1}{2n} \|X^t\theta - Y^t\|^2 + \sum_{i=1}^p \frac{\mu \gamma}{T} (\theta_i - P_{i.}\mathds 1 \log(\theta_i)) + \lambda \theta_i 
\end{equation}

Which can be rewritten as:

\begin{equation}
\label{eq:mtw_coef_g}
\min_{\theta \in \bbR^p_{++} } \frac{1}{2n} \|X^t\theta - Y^t\|^2 + \alpha \sum_{i=1}^p g_i(\theta_i)
\end{equation}

Where $g_i: x \mapsto  (x - a_i \log(x)) + b x$ with $\alpha = \frac{\mu \gamma}{T}$, $a = P\mathds 1$ and $b = \frac{\lambda T}{\gamma \mu}$. 

Computing the proximal operator of $G = \sum_i{g_i}$ boils down to carrying out the proximal operators $\prox_{\alpha g_i}$, element-wise.
Therefore, problem \eqref{eq:mtw_coef_g} can be solved using proximal coordinate descent \cite{fercoq} (Algorithm \ref{alg:cd}).

\begin{algorithm}[tb]
    \caption{Proximal coordinate descent}
    \label{alg:cd}
    \begin{algorithmic}
        \STATE {\bfseries Input:}  $X^t, Y^t, \alpha, P,$ descent steps $\eta_j = \frac{1}{\sum_{i=1}^n {X^t}_{ij}^2}$
        \STATE Initialize for $\theta = \theta_0$
        \REPEAT
        \FOR{$j=1$ {\bfseries to} $p$}
        \STATE$ \theta_j = \prox_{\alpha g_j}\left(\theta_j - \eta {X_{.j}^{t}}^\top(X^t - Y^t)\right)$
        \ENDFOR
        \UNTIL{convergence}
    \end{algorithmic}
\end{algorithm}

\paragraph{Sinkhorn's algorithm.}
The generalized Sinkhorn algorithm used to compute the Unbalanced Wasserstein barycenter may suffer from numerical as instability as the entropy regularization goes to zero i.e when $\epsilon \to 0$.
As recalled in Algorithm \ref{alg:sinkhorn}, the barycenter update requires taking the power $\frac{\gamma + \epsilon}{\epsilon}$ of the transport marginals. Typically for the value of $\epsilon = \frac{1}{mp}$ where $m$ is the median value of the cost matrix M, we encounter overflow errors for a certain range of hyperparameters. To allievate this problem, we rely on the log-stabilized version first introduced by \cite{schmitzer16}. Consider the change of variables $u'= u' \exp(a), v'= v' \exp(b)$. The idea is to absorbe the large values of the scaling variables in log-domain (i.e $a$ and $b$) while keeping $u'$ and $v'$ close to 1 as possible. We rely on this trick and allow our model to automatically switch to log-stabilized Sinkhorn when numerical errors are met.

\begin{algorithm}[b]
   \caption{Generalized Sinkhorn \cite{chizat:17}}
   \label{alg:sinkhorn}
   \begin{algorithmic}
       \STATE {\bfseries Input:}  $ \theta^1, \dots, \theta^T$
       \STATE Initialize for $(t = 1, \dots, T) \, (u^t, v^t) = (\mathds 1, \mathds 1)$, 
       \REPEAT
       \FOR{$t=1$ {\bfseries to} $T$}
       \STATE $u^t \gets \left(\frac{\theta^t}{Kv^t}\right)^{\frac{\gamma}{\gamma + \varepsilon}}$
       \ENDFOR
       \STATE $\thetabar \gets \left( \frac{\sum_{t=1}^T ( K^\top u^t) ^{ \frac{\varepsilon}{\varepsilon + \gamma} }}{T}\right)^{\frac{\varepsilon + \gamma}{\varepsilon}} $
       \FOR{$t=1$ {\bfseries to} $T$}
       \STATE $v^t \gets \left(\frac{\thetabar}{K^\top u^t}\right)^{\frac{\gamma}{\gamma + \varepsilon}}$
       \ENDFOR
       \UNTIL{convergence}
   \end{algorithmic}
\end{algorithm}
For simulations with synthetic images, we apply the Kernel matrix $\exp(-M / \epsilon)$ using fast convolutions which reduces considerably the complexity of the algorithm \cite{solomon:15}. Indeed, since our cost matrix $M$ is simply a separable euclidean distance over a square grid, applying the Kernel $K$ to an image is equivalent to computing convolutions its rows and then the columns of the obtained image. Moreover, this kernel separability property still be exploited in log-domain \cite{schmitz:17}.

\begin{figure}[H]
	\centering
	\includegraphics[width=\linewidth]{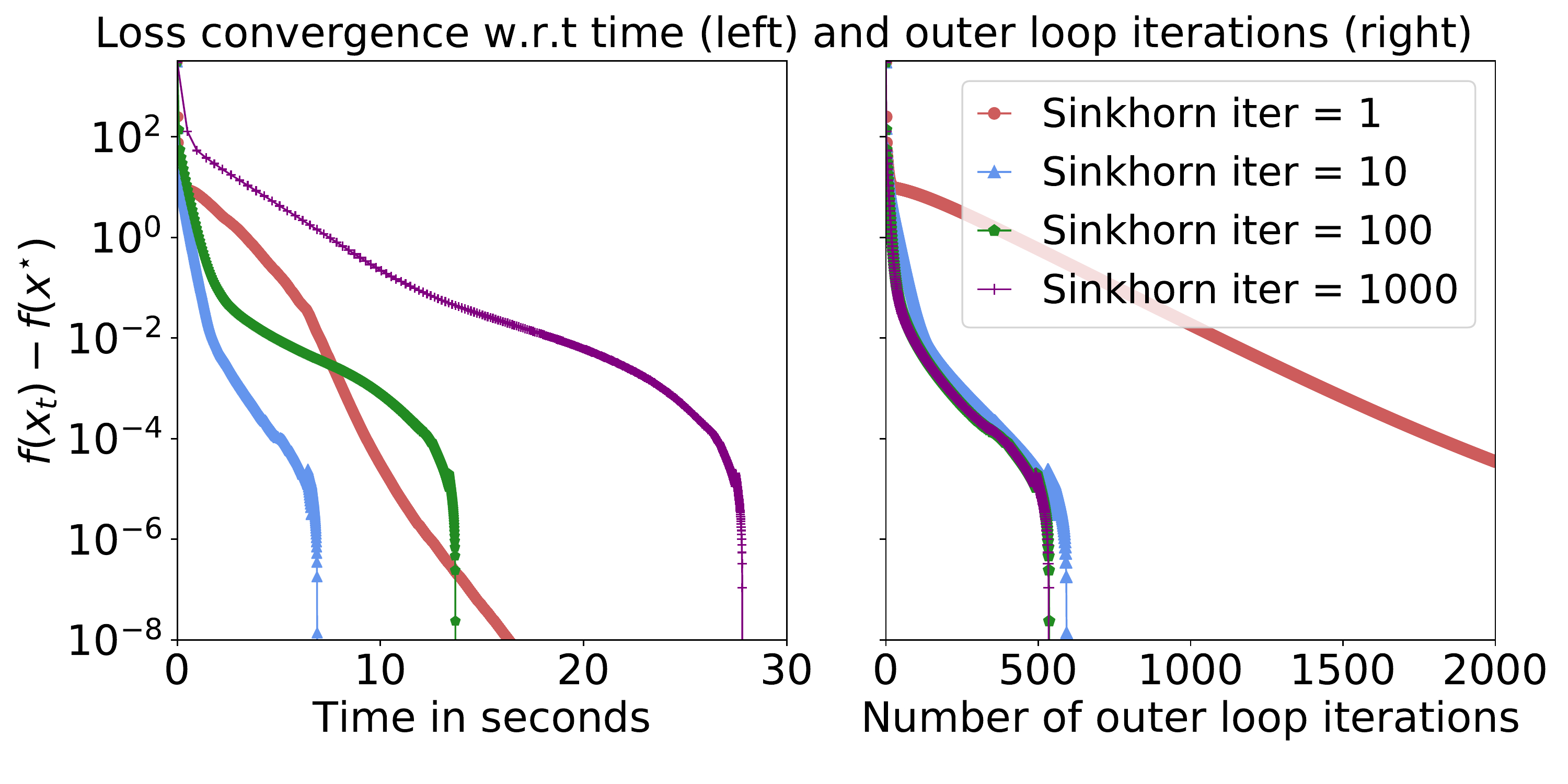}
	\caption{Illustration of alternating optimization trade-off. \label{f:iterot}}
\end{figure}

\paragraph{Alternating optimization.}
As discussed in section \ref{s:optim}, the minimized loss is jointly convex. We observe that in practice, performing a few tens of iterations of Sinkhorn speeds up the convergence. This trade-off  is illustrated in Figure \ref{f:iterot} where we show the optimality gap of the loss function w.r.t to different numbers of iterations of Sinkhorn updates. For proximal coordinate descent however, we wait for convergence in each inner loop.

\section{Dirty models}
\label{s:dirty}
In this section we show that for Dirty models, hyperparameters need not to be tuned over a 2D grid but within a surface between the lines with slopes 1 and $\frac{1}{\sqrt{T}}$ where T is the number of tasks.
Recall the optimization solved by Multi-task Dirty models with $\ell_1/\ell_2$ norms:
\begin{equation}
    \label{eq:dirty}
    \min_{\substack{\theta^1, \theta^2 \\ \in \bbR^{p \times T}}} \sum_{t=1}^T \frac{1}{2n} \| X^t \theta_c^t + X^t \theta_s^t - Y^t \|^2 + \mu \|\Theta_c\|_{2,1} + \lambda \|\Theta_s\|_1   \enspace ,
   \end{equation}

Let's denote the column stacking $\Theta = \left[\theta^1, \dots,
 \theta^T\right]$ and similarly the block diagonal matrix $\boldsymbol{X} = \diag{X^1, \dots, X^T}$ and $\boldsymbol{Y} = \boldsymbol{X}\Theta$.

The optimality condition for problem \eqref{eq:dirty} reads:
\begin{equation*}
\label{eq:fermat}
0 \in \boldsymbol{X}^\top(\boldsymbol{X}\Theta^*_c + \boldsymbol{X}\Theta^*_s  - \boldsymbol{Y}) + \mu \partial_{\ell_{21}}(\Theta^*_c) +
\lambda \partial_{\ell_{1}}(\Theta^*_s)
\end{equation*}

The subdifferential of $\ell_{21}$ is simply the projection over the unit ball of its dual norm $\ell_{2\infty}$ at $\Theta \neq 0$ and is the set of all elements of that ball otherwise. Thus, for $\Theta^*$ equal to $0$ we get:
\begin{align*}
\|\boldsymbol{X}^\top \boldsymbol{Y}\|_{2\infty} \leq \mu \\
\|\boldsymbol{X}^\top \boldsymbol{Y}\|_{\infty} \leq \lambda
\end{align*}

The bounds above define a rectangular box over which the gridsearch must be performed. However, we can show that this gridsearch can be reduced to a much smaller triangle.

Suppose $\exists (j, k)$ s.t $\Theta_s^{j, k} \neq 0$. Therefore 
\begin{align*}
\exists Z_c \in & \mu \partial_{\ell_{21}}(\Theta^*_c) \quad  \mu |Z_c^{j, k} | = \lambda \\ 
&\Rightarrow \quad \mu \geq \lambda
\end{align*}

Thus, when $\lambda > \mu$, the model reduces to an independent Lasso estimator.

Now suppose $\exists (j, k)$ s.t. $\Theta_c^{j, k} \neq 0$. Therefore 
\begin{align*}
\exists Z_s \in & \mu \partial_{\ell_{1}}(\Theta^*_s) \quad  \mu \frac{\Theta_s^{j, k}}{\| \Theta_s^j \|_2} = \lambda Z_s^{j, k}  \\ 
&\Rightarrow \quad \mu \leq \sqrt{T} \lambda
\end{align*}

Thus, when $\sqrt{T} \lambda < \mu$, the model reduces to a group-Lasso estimator.

\section{Simulation details}

\begin{figure}[H]
    \centering
        \includegraphics[width=0.8\linewidth]{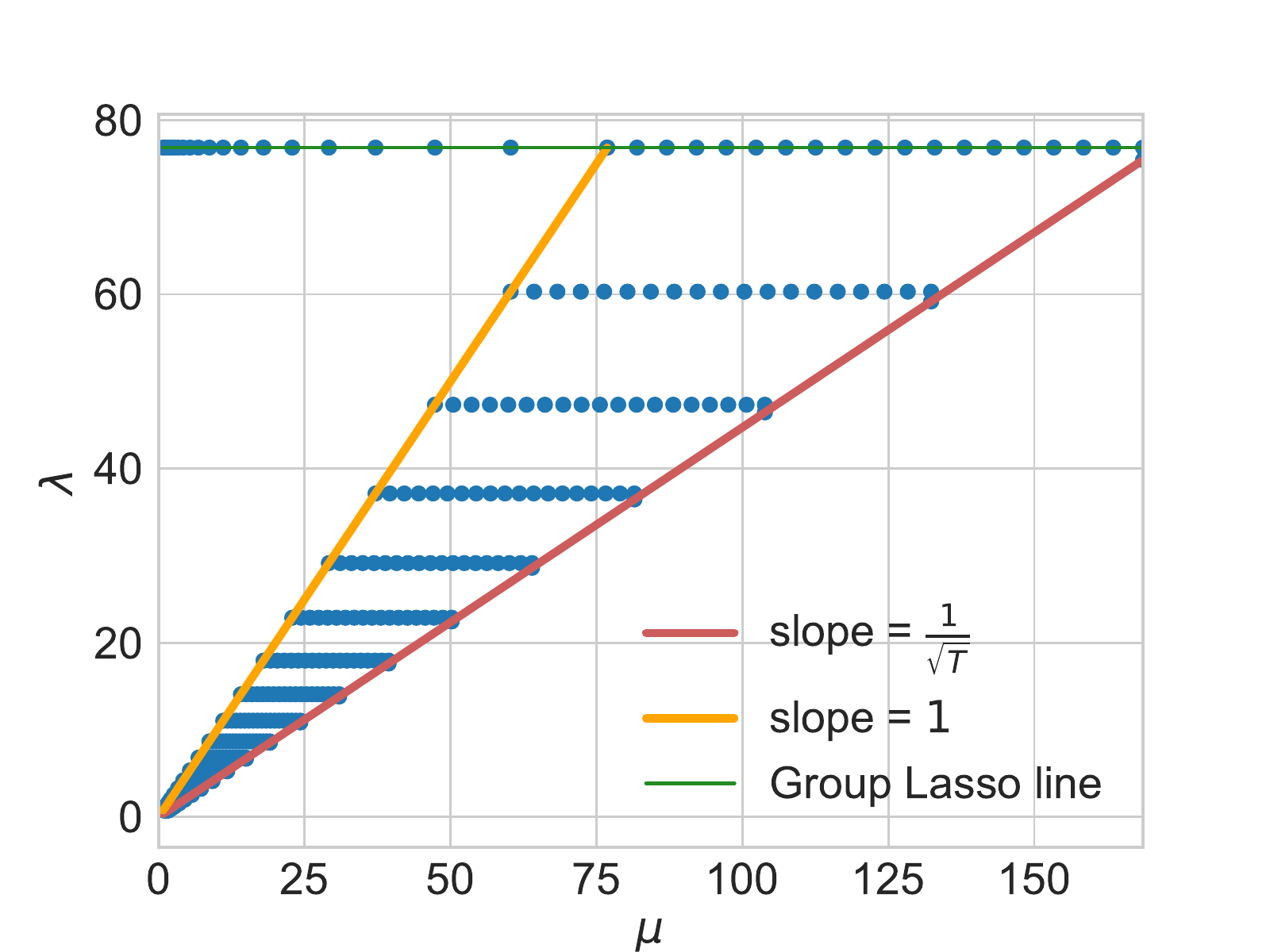}
        \caption{Illustration of a hyperparamers grid sampling for Dirty models. \label{f:grid}}
\end{figure}

\paragraph{model selection.}
\label{s:expes}
For all simulations, we selected the best hyperparameters of each model among a set of hyperparameters set as follows.
For Lasso, we set a logarithmic scale of 100 values between  $\mu_{\text{max}} = \|\boldsymbol{X}^{\top} Y \|_{\infty}$ and $ \frac{\mu_{\text{max}}}{100}$. The tuning grid of Dirty models is given in section{s:dirty}. In practice we start by sampling 15 points on the base of the triangle that we further divide by a logarithmic sequence between $\lambda_{\text{max}} =  \|\boldsymbol{X}^{\top} Y \|_{2\infty}$ and $\frac{\lambda_{\text{max}}}{100}$. Moreover, we sample 20 points over the line $y = \mu_{\text{max}}$ for exclusive Group Lasso models. Figure \ref{f:grid} shows an illustrative example of the sampled hyparaparameters.

For MTW, $\mu$ is chosen among 10 candidates within a logarithmic scale between 1 and 100. The list of 20 values of $\lambda$ is the same as the one used for the independent Lasso models.

\begin{figure}[h]
	\centering
	\adjincludegraphics[trim={6cm 12cm 6cm 12cm},clip, width=0.8\linewidth]{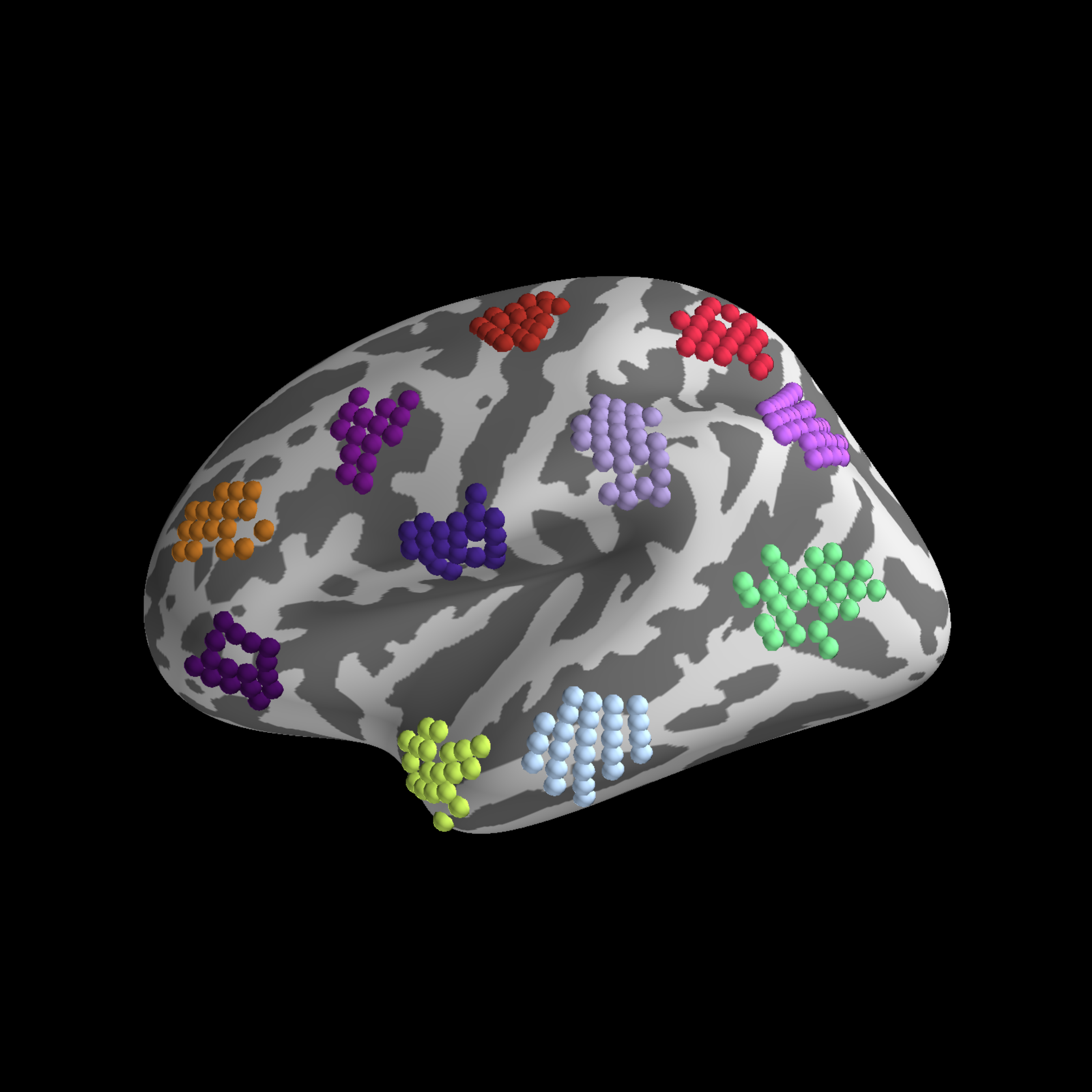}
	\caption{Areas from which non-zero features are selected.
		\label{f:brainregions}}
\end{figure}

\paragraph{MEG source localization}
The supports of the simulated brain activations (regression coefficients) are selected by taking one non-zero feature in each region illustrated in Figure \ref{f:brainregions}. If a regression coefficient is $k$-sparse, $k$ regions are selected in which one random feature is non-zero.

\begin{figure}[t]
    \begin{minipage}{0.49\linewidth}
	\includegraphics[width=\linewidth]{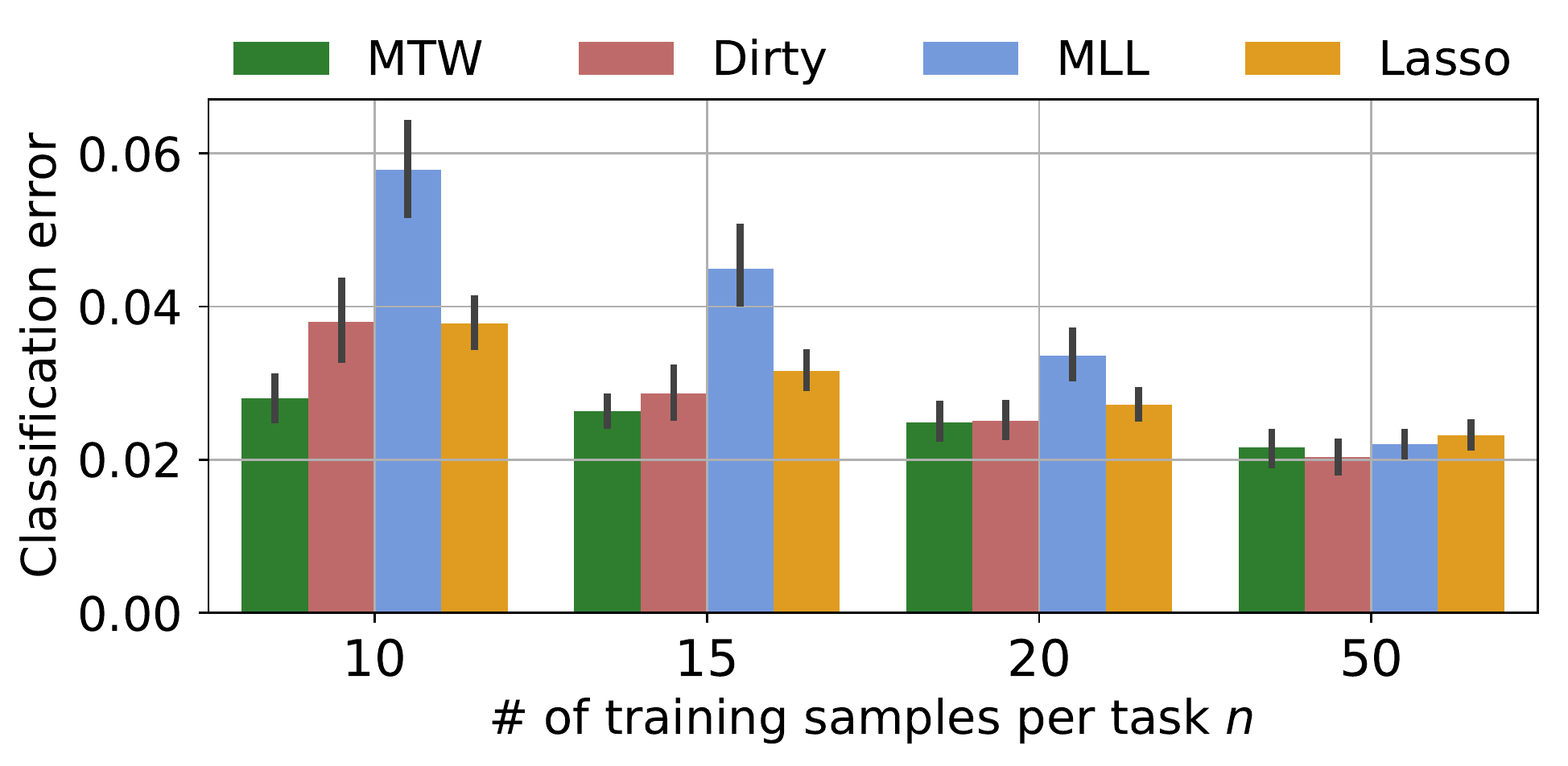}
	(Digit 0)
	\end{minipage}
    \begin{minipage}{0.49\linewidth}
	\includegraphics[width=\linewidth]{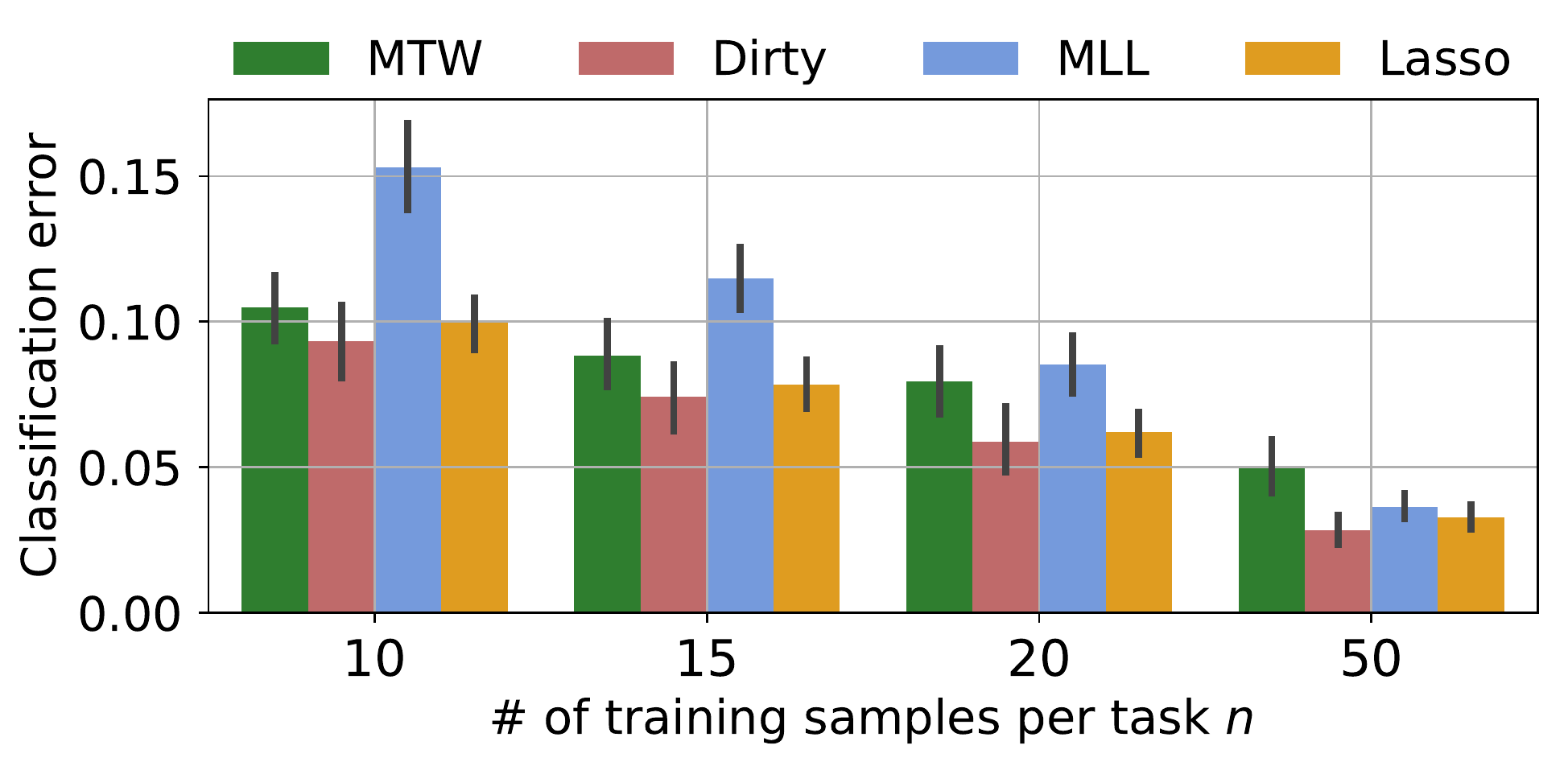}
	(Digit 1)
	\end{minipage}

    \begin{minipage}{0.49\linewidth}
	\includegraphics[width=\linewidth]{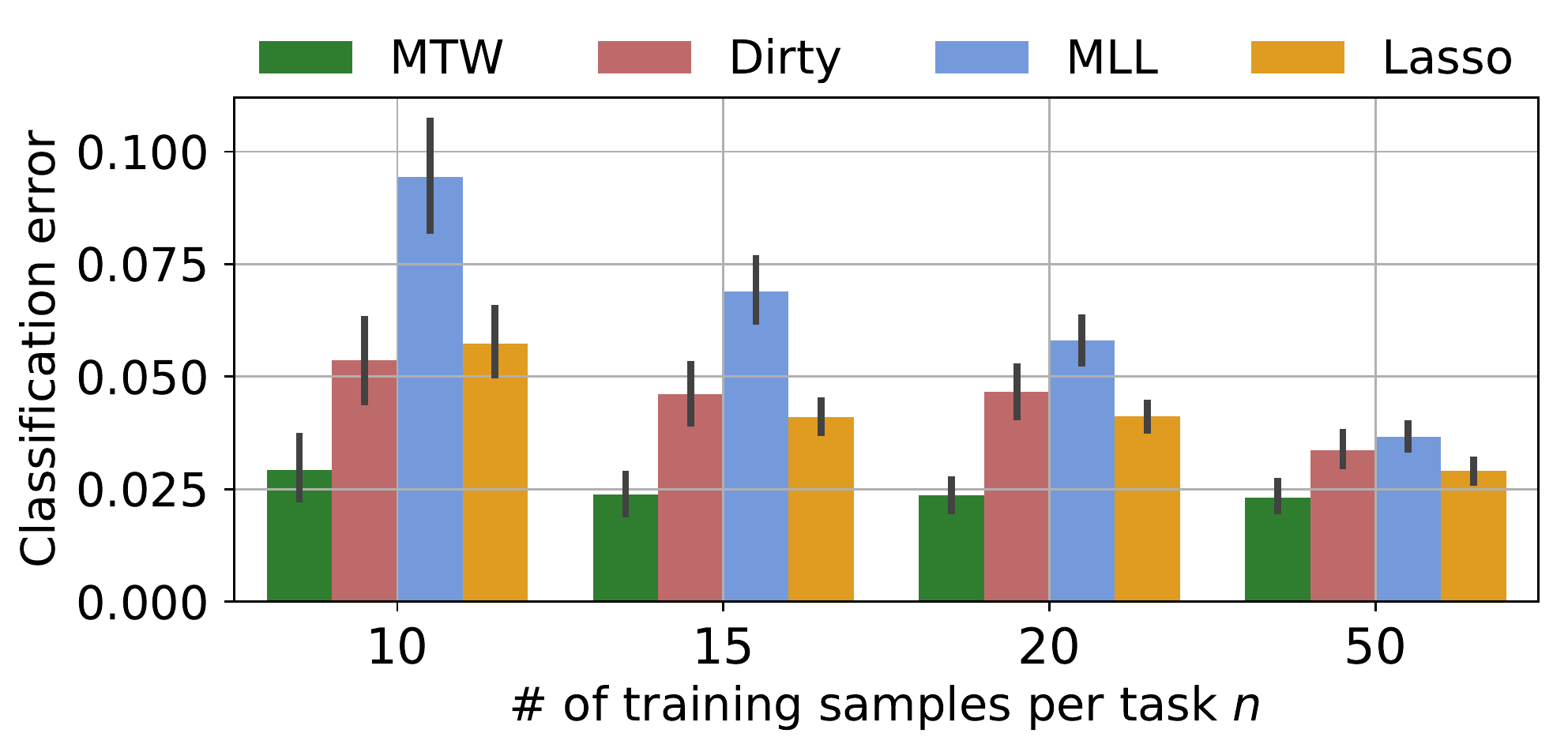}
	(Digit 2)
	\end{minipage}
    \begin{minipage}{0.49\linewidth}
	\includegraphics[width=\linewidth]{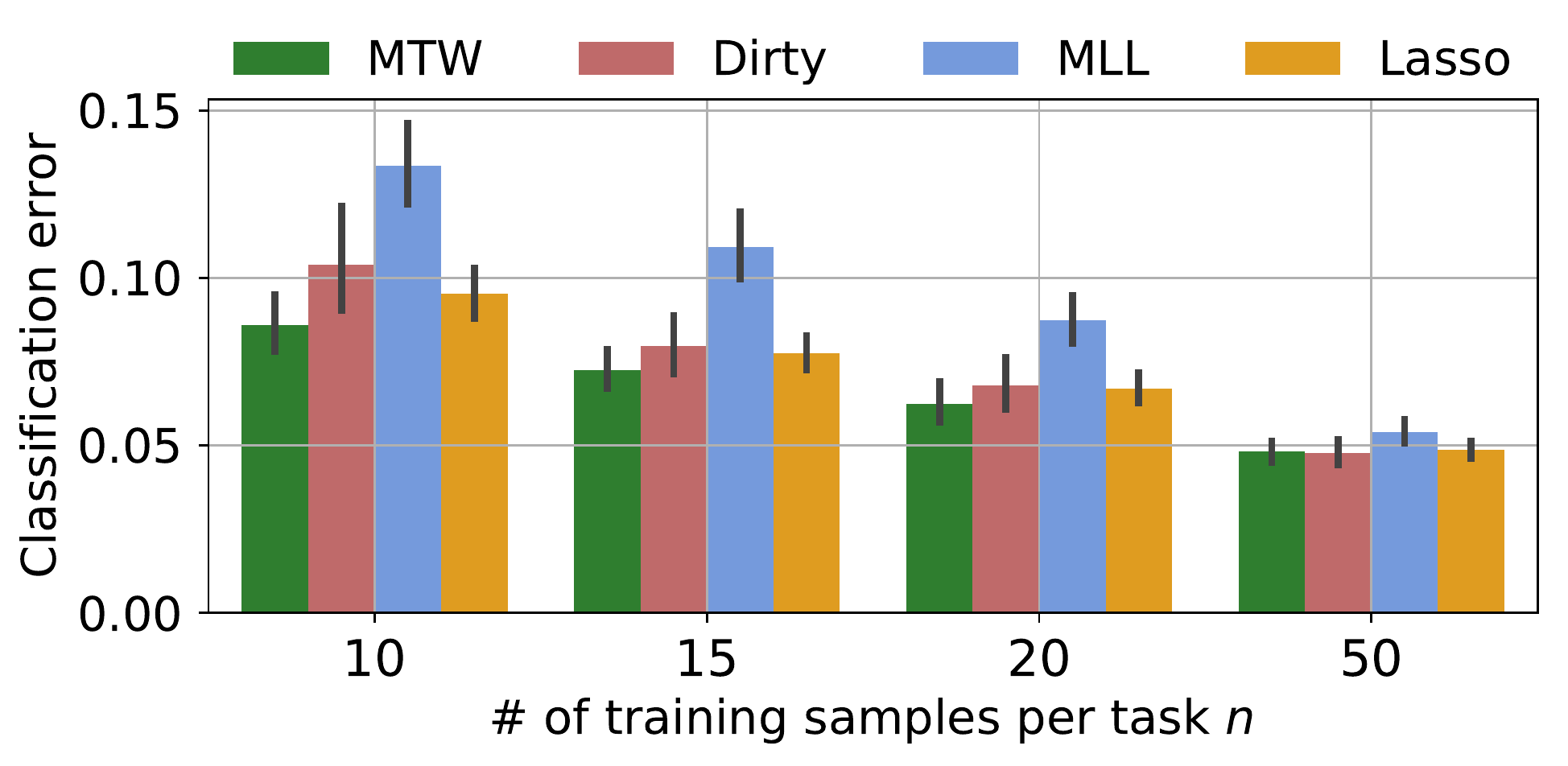}
	(Digit 3)
	\end{minipage}

    \begin{minipage}{0.49\linewidth}
	\includegraphics[width=\linewidth]{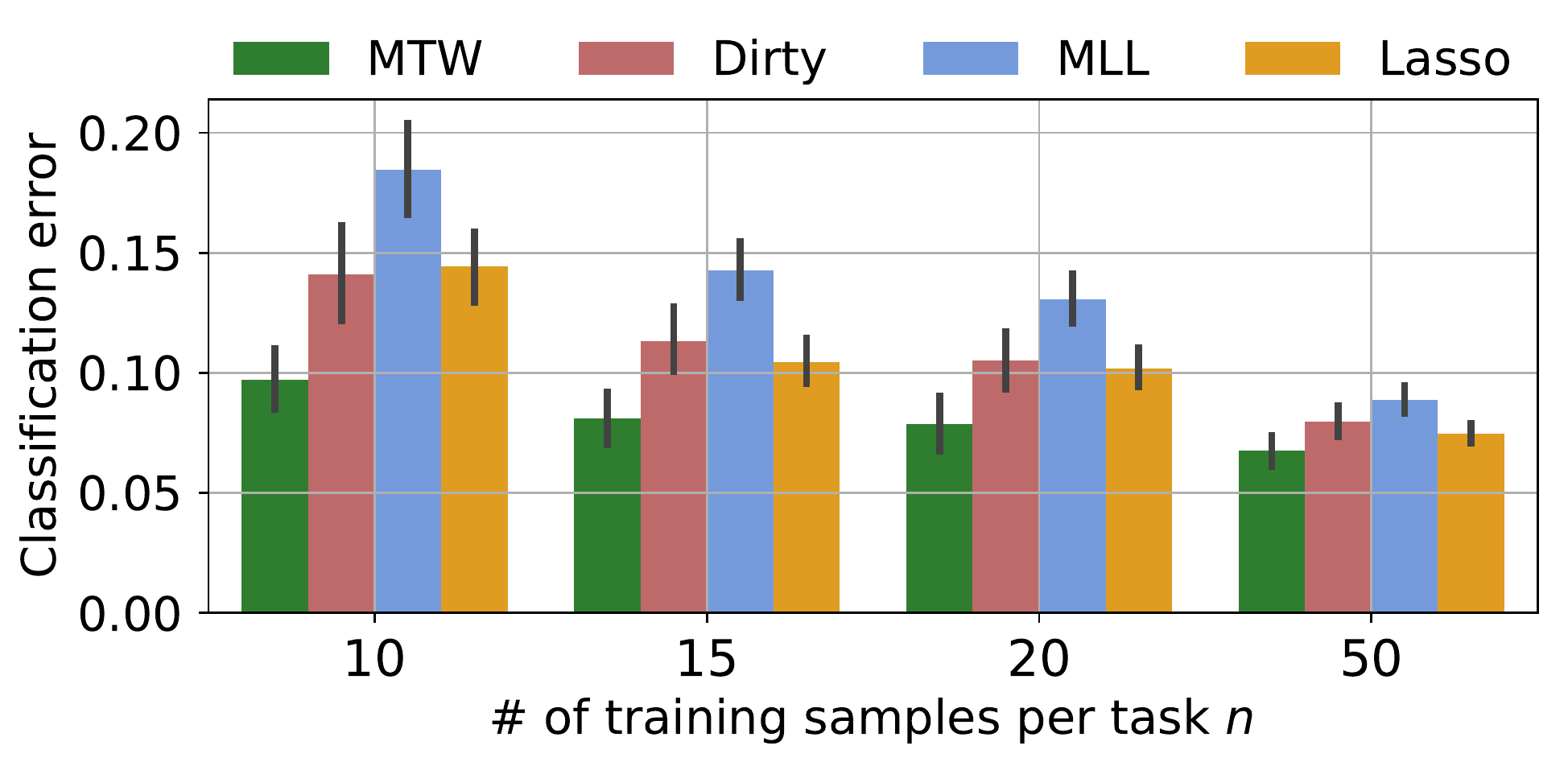}
	(Digit 4)
	\end{minipage}
    \begin{minipage}{0.49\linewidth}
	\includegraphics[width=\linewidth]{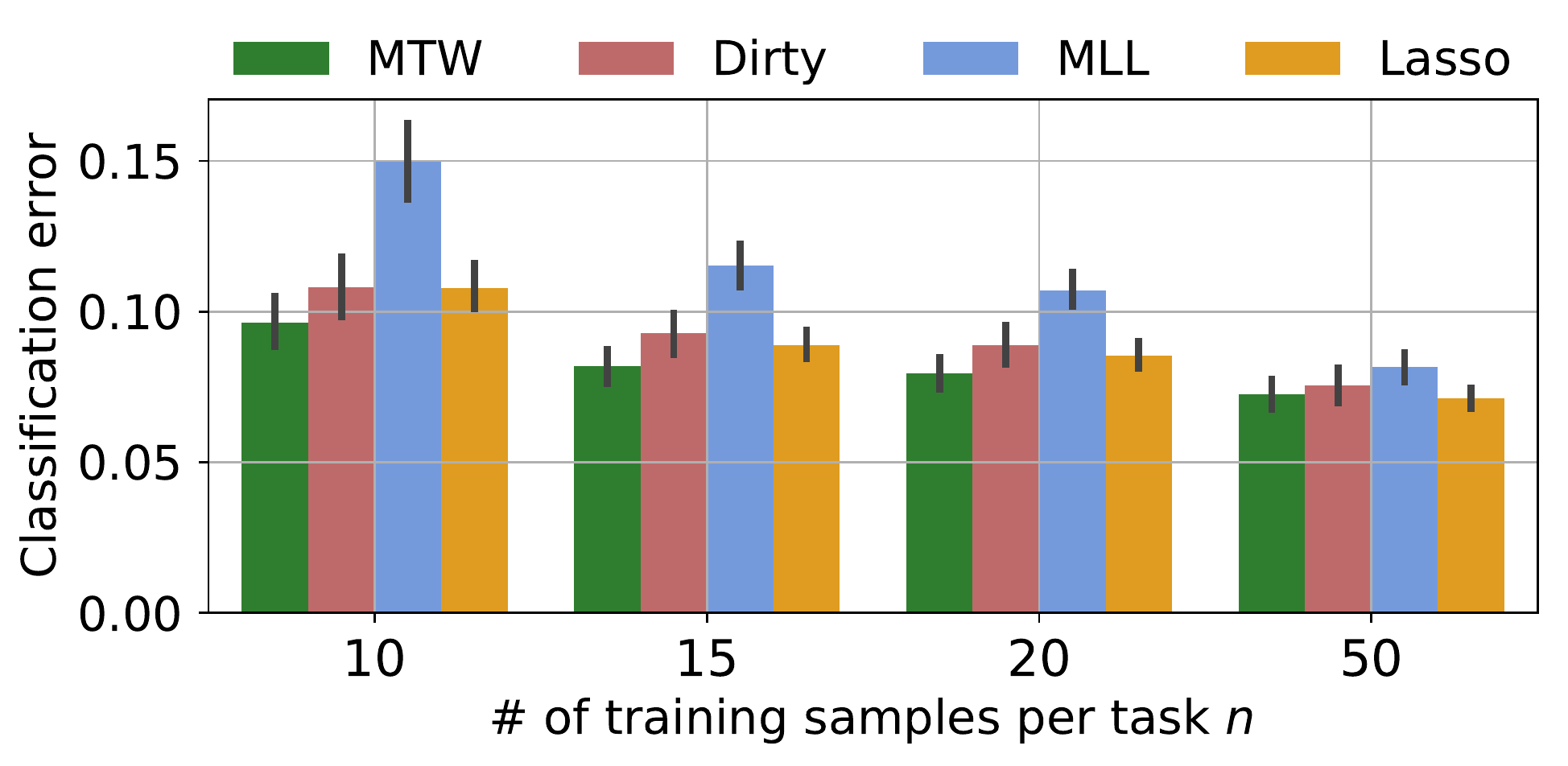}
	(Digit 5)
	\end{minipage}

	\caption{Mean classification error per task (digit in (`0'--`5').)
		\label{f:digitsresultssup}}
\end{figure}

\paragraph{Handwritten digits recognition.}
We concatenate the handwritten digits dataset of \cite{digits} as a matrix $X \in \bbR^{nt \times p}$ where we selected the 6 first tasks (corresponding to the 6 first numerals 0-5) i.e $T=6$; and the number of features $p=240$ corresponding to 15 $\times$ 16 reduced images. The number of samples per task $n$ is set to 10; 15; 20 and 50. We concatenate the one-hot encoded binary vector for each task $Y^t \in \bbR^{nT}$ so as to perforum one versus all classification. Thus, $X$ is the design matrix common to all regression tasks. For each task, the dataset contains 200 samples. Model selection if performed by first isolating a validation set of 200 - $n$ samples per task. And computing a 5-fols Cross-validation error score on the training set. We performed 20 random selections of the validation samples and reported the mean classification errors in Figure \ref{f:digitsresults}. The detailed classification errors per task (taking the mean only across randomized splits) are displayed in Figure \ref{f:digitsresultssup}. 
\begin{figure}[t]
	\centering
	\includegraphics[width=\linewidth]{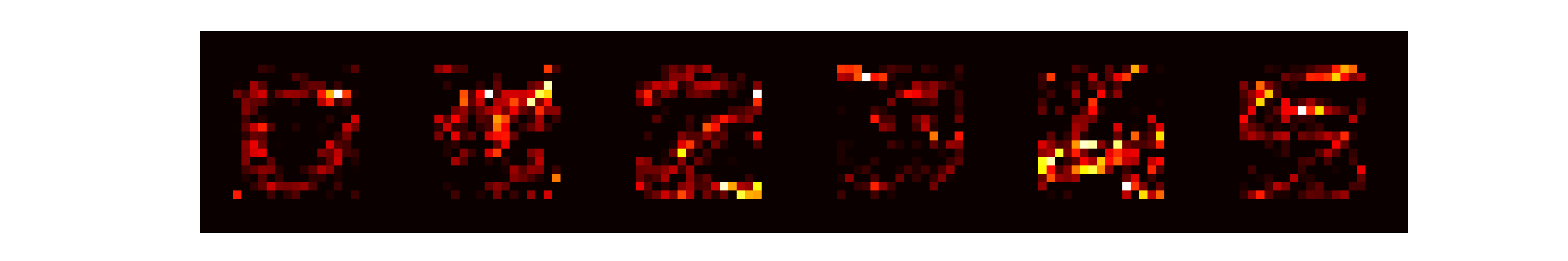}
	\centering 
	(MTW)

	\adjincludegraphics[,clip, width=\linewidth]{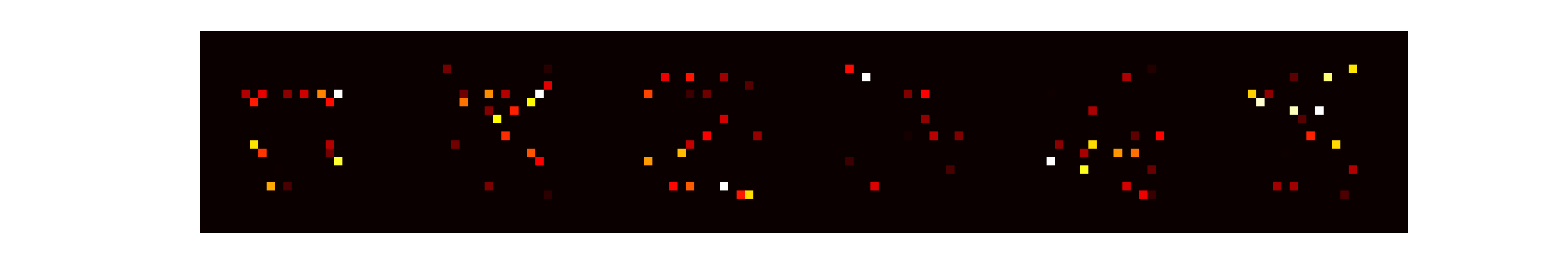}
	\centering
	(MLL)
	
	\centering
\includegraphics[width=\linewidth]{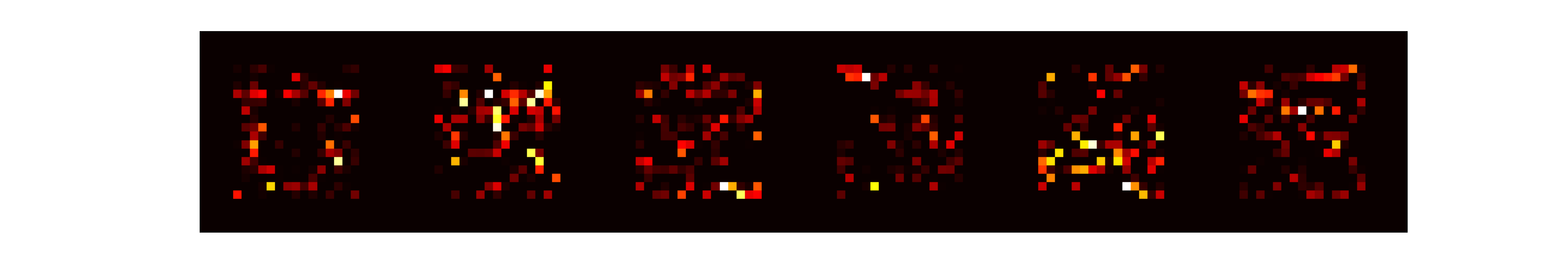}
\centering 
(Dirty)

	\centering
\includegraphics[width=\linewidth]{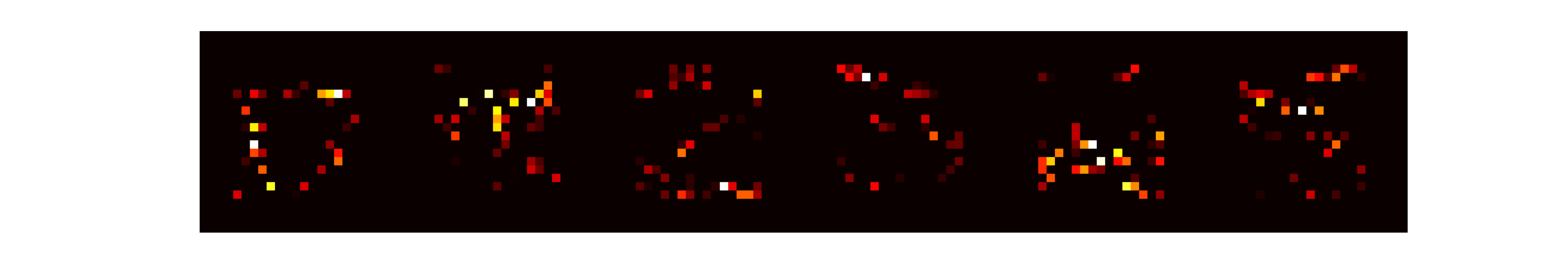}
\centering 
(Lasso)

	\caption{Learned regression coefficients $\bs\theta_+$ corresponding to the digits (`0'--`5').
		\label{f:alldigitscoefs}}
\end{figure}
We display in Figure \ref{f:alldigitscoefs} the learned regression coefficients by all methods.
\onecolumn

\section{Python code}
\label{s:code}
\paragraph{Alternating Optimization.}

\begin{python}
def solver_mtw(X, Y, theta01=None, theta02=None, mu=1., lambda_=0., M=None,
               epsilon=0.01, gamma=1., stable=False, maxiter=2000,
               callback=None, tol=1e-5, maxiter_ot=20, tol_ot=1e-4,
               positive=False, returnlog=True, R=None):
    """Perform Alternating Optimization of the MTW problem.

    Parameters
    ----------
    X : numpy array (n_tasks, n_features, n_samples)
    Y : numpy array (n_tasks, n_samples)

    theta01 : numpy array (n_features, n_tasks)
        initial positive parts.
    theta02 : numpy array (n_features, n_tasks)
        initial negative parts.
    mu: float >= 0.
        OT regularization hyperparameter.
    lambda_: float >= 0.
        L1 penalty regularization hyperparameter.
    M: numpy array (n_features, n_features)
        OT Ground metric.
    maxiter : int > 0. optional, default 2000
        maximum number of alternating iterations.
    positive: bool. optional.
        If True, coefficients are constrained to be non-negative.
    callback : callable. optional, default None.
        printing function.
    tol : float > 0. optional, default 1e-5
        Stopping criterion threshold on relative loss decrease.
    tol_ot : float > 0. optional, default 1e-4
        Stopping criterion threshold of Sinhorn.
    R: numpy array (n_tasks, n_samples)
        regression residuals for warm-start.
    returnlog : boolean. optional, default False
        if True, returns convergence log.

    Returns
    -------
    theta : numpy array (n_features, n_tasks)
        optimal minimizer

    if `returnlog` == True:
        theta : numpy array
            optimal minimizer
        log : dict.
            objectives, errors.

    """
    log = {'loss': [], 'dloss': [], 'log_sinkhorn1': [], 'log_sinkhorn2': []}
    n_tasks, n_samples, n_features = X.shape
    if theta01 is None:
        coefs01 = np.ones((n_features, n_tasks)) / n_features
    if theta02 is None:
        coefs02 = np.ones((n_features, n_tasks)) / n_features

    marginals1 = np.ones((n_tasks, n_features)) / n_features
    marginals2 = np.ones((n_tasks, n_features)) / n_features

    Xf = np.asfortranarray(X)  # fortran order for numba
    Yf = np.asfortranarray(Y)
    theta1 = coefs01.copy()
    theta2 = coefs02.copy()
    theta = theta1 - theta2

    thetaold = theta.copy()
    Ls = lipschitz_numba(np.asfortranarray(X))
    Ls[Ls == 0.] = Ls[Ls != 0.].min()

    # If inputs are images, then use Kernal convolutions in Sinkhorn
    ot_img = True
    if len(M) == n_features:
        ot_img = False

    update_ot_1 = set_ot_func(stable, ot_img)
    update_ot_2 = set_ot_func(stable, ot_img)

    t_cd = 0.
    t_ot = 0.
    xp = get_module(M)
    K = xp.exp(- M / epsilon)

    # Initial barycenters (positive, negative parts)
    thetabar1 = np.ones_like(coefs01).mean(axis=-1)
    thetabar2 = np.ones_like(coefs02).mean(axis=-1)
    thetabar = thetabar1 - thetabar2

    # if non-nenegativity constraint, negative parts = 0
    if positive:
        theta2 *= 0.
        thetabar2 *= 0.
        theta = theta1.copy()

    # Begin alternting optimization loop
    for i in range(maxiter):
        marginals1 = np.asfortranarray(marginals1)

        t = time()
        if not positive:
            theta2f = np.asfortranarray(theta2)
            Y1 = utils.residual(Xf, - theta2f, Yf)  # compute Yf + Xf.dot(theta2f) 
        else:
            Y1 = Yf

        # Do proximal coordinate descent to update theta 1
        theta1, R, obj = update_coefs(Xf, Y1, Ls, marginals1,
                                      coefs0=theta1,
                                      R=R,
                                      mu=mu,
                                      gamma=gamma,
                                      lambda_=lambda_,
                                      tol=1e-6,
                                      maxiter=10000)
        if not positive:
            theta1f = np.asfortranarray(theta1)
            marginals2 = np.asfortranarray(marginals2)

            Y2 = utils.residual(Xf, theta1f, Yf)  # compute Yf - Xf.dot(theta1f) 
            theta2, R, obj = update_coefs(- Xf, Y2, Ls, marginals2,
                                          coefs0=theta2,
                                          R=R,
                                          mu=mu,
                                          gamma=gamma,
                                          lambda_=lambda_,
                                          tol=1e-6,
                                          maxiter=10000)
            theta = theta1 - theta2
            obj += lambda_ * theta1.sum()
        else:
            theta = theta1.copy()

        t_cd += time() - t
        dx = abs(theta - thetaold).max() / max(1, thetaold.max(), theta.max())

        thetaold = theta.copy()

        # move thetas to gpu for Sikhorn
        theta1_gpu = xp.asarray(theta1)
        theta2_gpu = xp.asarray(theta2)

        t = time()

        # compute barycenters
        if mu:
            fot1, log_ot1, marginals1, u1, bar1 = update_ot_1(theta1_gpu, M,
                                                            epsilon,
                                                            gamma,
                                                            K=K,
                                                            tol=tol_ot,
                                                            maxiter=maxiter_ot)
            # If unstable, move to log domain computations
            if fot1 is None:
                warnings.warn("""Nan found in positive, re-fit in log-domain.""")
                u1 = np.log(u1 + 1e-100)  # move scaling u to log domain
                stable = True
                update_ot_1 = set_ot_func(True, ot_img)
                fot1, log_ot1, marginals1, bar2 = \
                    update_ot_1(theta1_gpu, M, epsilon, gamma, K=K, u=u1,
                                tol=tol_ot, maxiter=maxiter_ot)
            log["log_sinkhorn1"].append(log_ot1["cstr"])
            thetabar1 = bar1
            obj += mu * fot1 / n_tasks

            if not positive:
                fot2, log_ot2, marginals2, u2, bar2 = \
                    update_ot_2(theta2_gpu, M, epsilon, gamma, K=K,
                                tol=tol_ot, maxiter=maxiter_ot)
                if fot2 is None:
                    warnings.warn("""Nan found in negative, re-fit in log-domain.""")
                    u2 = np.log(u2 + 1e-100)
                    stable = True
                    update_ot_2 = set_ot_func(True, ot_img)
                    fot2, log_ot2, marginals2, u2, bar2 = \
                        update_ot_2(theta2_gpu, M, epsilon, gamma, K=K, u=u2,
                                    tol=tol_ot, maxiter=maxiter_ot)

                log["log_sinkhorn2"].append(log_ot2["cstr"])
                thetabar2 = bar2
                obj += mu * fot2 / n_tasks
                thetabar = thetabar1 - thetabar2
            else:
                thetabar = thetabar1

        t_ot += time() - t
        if callback:
            callback(theta, thetabar, v=obj)

        log['loss'].append(obj)
        log['dloss'].append(dx)

        # dx < tol:
        if dx < tol:
            break
    if i == maxiter - 1:
        print("\n"
              "******** WARNING: Stopped early in main loop. *****\n"
              "\n"
              "You may want to increase mtw.maxiter.")

    if callback:
        print("Time ot 

    log['stable'] = stable
    if positive:
        theta2 *= 0.
        thetabar2 = np.zeros_like(thetabar1)
        marginals2 = np.zeros_like(marginals1)
        u2 = np.ones_like(u2)
    if returnlog:
        return theta, thetabar, log
    return theta, thetabar
\end{python}

\paragraph{Generalized Sinkhorn}
\begin{python}

def barycenterkl(P, M, epsilon, gamma, K=None, u=None, tol=1e-4,
                 maxiter=1000):
    """Compute Unblanced Wasserstein barycenter.
    P: numpy array (n_features, n_tasks)
        positive regression coefficients.
    M: numpy array (n_features, n_featuresq)
        Ground metric
    epsilon: float  > 0
        Entropy hyperparameter
    gamma: float > 0
        KL marginals hyperparameter
    K: numpy array (n_features, n_features)
        exp(- M / epsilon)
    u: numpy array (n_features, n_tasks)
        scaling vector for warm-start
    """
    xp = get_module(P)
    frac = gamma / (gamma + epsilon)
    n_features, n_tasks = P.shape
    frac = gamma / (gamma + epsilon)
    if u is None:
        u = xp.ones((n_features, n_tasks))
    Ku = K.dot(u)

    log = {'cstr': [], 'flag': 0, 'obj': []}
    weights = xp.ones(n_tasks) / n_tasks
    q = xp.ones(n_features)
    qold = q.copy()
    return_nan = False
    for i in range(maxiter):
        a = (P / Ku) ** frac
        Ka = K.T.dot(a)
        q = ((Ka  ** (1 - frac)).dot(weights))
        q = q ** (1 / (1 - frac))
        Q = q[:, None]
        cstr = abs(q - qold).max() / max(q.max(), qold.max(), 1)
        qold = q.copy()
        u_old = u.copy()
        u = (Q / Ka) ** frac

        # If stability problems, return nan to switch to log in alg1
        if not xp.isfinite(u).all():
            return_nan = True
            break
        Ku = K.dot(u)
        log["cstr"].append(cstr)
        if abs(cstr) < tol:
            break

    if i == maxiter - 1:
        warnings.warn("Early stop, Maxiter too low !")
        log['flag'] = - 1
    marginals = (a * Ku).T

    try:
        marginals = marginals.get()
        u = u.get()  # Move back to CPU
        u_old = u_old.get()
        q = q.get()
    except AttributeError:
        pass

    # compute loss
    f = utils.wklobjective_converged(P, n_tasks * q, n_tasks * 0,
                                     marginals.sum(),
                                     epsilon, gamma)
    if return_nan or np.isnan(f):
        f = None
        u = u_old
    return f, log, marginals, u, q

\end{python}

\appendix
\numberwithin{figure}{section}
\numberwithin{equation}{section}

\subfilebiblio{}